\documentclass[a4paper]{article}

\usepackage{graphicx}

\usepackage{mathptmx}      

\usepackage{todonotes}
\usepackage{url}

\usepackage{a4wide}

\usepackage[authoryear,round,colon,sectionbib]{natbib}

\usepackage{amsmath}
\usepackage{amssymb}
\usepackage{tipa}
\usepackage{fancyvrb}
\usepackage{xfrac}
\usepackage{subfig}
\usepackage{multirow}
\usepackage{booktabs}
\usepackage{comment}

\newtheorem{definition}{Definition}

\renewcommand{\cite}[1]{\citep{#1}}

\newcommand{\hreg}{\textrm{h}}
\newcommand{\hinv}{\textrm{\textturnh}}

\newcommand{\data}[1]{\textsl{#1}}
\newcommand{\alg}[1]{\textsc{#1}}

\newcommand{\mdl}{\ensuremath{\mathbf{m}}}

\begin{document}

\title{On Cognitive Preferences and \\ the Plausibility of Rule-based Models
}




  \author{ 
    Johannes F\"urnkranz \\
    TU Darmstadt\\
    Department of Computer Science\\
    Hochschulstra{\ss}e 10 \\
    D-64289 Darmstadt, Germany\\
    \url{juffi@ke.tu-darmstadt.de}\\ 
    \and
  Tom\'{a}\v{s} Kliegr \\
    University of Economics, Prague\\
    Department of Information and Knowledge Engineering \\
    n{\'a}m Winstona Churchilla  4 \\
    13067 Prague, Czech Republic \\
    \url{tomas.kliegr@vse.cz}
    \and
    Heiko Paulheim \\
    University of Mannheim \\
    Institut f\"ur Informatik und Wirtschaftsinformatik\\
    D-68159 Mannheim, Germany\\
    \url{heiko@informatik.uni-mannheim.de}
}

\date{V4.0, April 2019}

\maketitle

\begin{abstract}
It is conventional wisdom in machine learning and data mining that logical
models such as rule sets are more interpretable than other models,
and that among such rule-based models, simpler models are more
interpretable than more complex ones. In this position paper, we question
this latter assumption by focusing on one particular aspect of interpretability,
namely the plausibility of models. Roughly speaking, we equate the plausibility of a
model with the likeliness that a user accepts it as an explanation for a prediction.
In particular, we argue that---all other
things being equal---longer explanations may be more convincing than shorter
ones, and that the predominant bias for shorter models, which is typically 
necessary for learning powerful discriminative models, may not be suitable
when it comes to user acceptance of the learned models. To that end, we first recapitulate evidence for and against this
postulate, and then report the results of an evaluation in a crowdsourcing study based on about 3,000 judgments.
 The results do not reveal a
strong preference for simple rules, whereas we can observe
a weak preference for longer rules in some domains. 
We then 
relate these results to well-known cognitive biases such as the conjunction fallacy,
the representative heuristic, or the recogition heuristic, and investigate their relation to rule length and plausibility.
\end{abstract}


\newpage

\section{Introduction}
\label{sec:introduction}

\enlargethispage*{12pt}

In their classical definition of the field,
\citet{Fayyad:1996:KPE:240455.240464} have defined knowledge discovery
in databases as ``\emph{the non-trivial process of identifying valid,
  novel, potentially useful, and ultimately understandable patterns in
  data.}'' Research has since progressed considerably in all of these
dimensions in a mostly data-driven fashion. The validity of models is
typically addressed with predictive evaluation techniques such as
significance tests, hold-out sets, or cross validation
\citep{EvaluatingLearningAlgorithms}, techniques which are now also
increasingly used for pattern evaluation
\citep{DiscoveringSignificantPatterns}. The novelty of patterns is
typically assessed by comparing their local distribution to expected
values, in areas such as novelty detection
\citep{NoveltyDetection-1,NoveltyDetection-2}, where the goal is to
detect unusual behavior in time series, subgroup discovery
\citep{SupervisedDescriptiveRuleDiscovery}, which aims at discovering
groups of data that have unusual class distributions, or exceptional
model mining \citep{ExceptionalModelMining}, which generalizes this
notion to differences with respect to data models instead of data
distributions. 
The
search for useful patterns has mostly been addressed via optimization, where the utility of a pattern is defined
via a predefined objective function \citep{HighUtilityPatterns} or via
cost functions that steer the discovery process into the direction of
low-cost or high-utility solutions
\citep{CostSensitive-Foundations}. To that end,
\citet{dm:microeconomic} formulated a data mining framework based on
utility and decision theory.

Arguably, the last dimension, understandability or interpretability, has received the
least attention in the literature. 
The reason why interpretability has rarely been explicitly addressed
is that it is often equated with the presence of logical or structured models
such as decision trees or rule sets, which have been extensively researched
since the early days of machine learning. In fact, much of the research on
learning such models has been motivated with their interpretability.
For example, \citet{jf:Book-Nada} argue that rules ``\emph{offer the
  best trade-off between human and machine understandability}''.  
Similarly, it has been argued that rule induction offers a good ''mental fit'' to decision-making problems \cite{RuleInduction-IDA,MentalFit-DataFit}.
Their
main advantage is the simple logical structure of a rule, which can be
directly interpreted by experts not familiar with machine
learning or data mining concepts.  Moreover, rule-based models are highly modular,
in the sense that they may be viewed as a collection of local
patterns
\cite{jf:Dagstuhl-04,jf:LeGo-08-WS-Paper,jf:LeGo-SI-Editorial}, whose
individual interpretations are often easier to grasp than the complete
predictive theory. For example, 
\citet{InterpretableDecisionSets} argued that rule sets (which they call decision sets) are more interpretable than decision lists, because they can be decomposed into individual local patterns. 

Only recently, with the success of highly precise but largely inscrutable deep learning models, has the topic of interpretability received serious attention, 
and 
several workshops in various disciplines have been
devoted to the topic of learning interpretable models at conferences like ICML
\citep{WS-HumanInterpretability-ICML16,WS-HumanInterpretability-ICML17,WS-HumanInterpretability-ICML18}, NIPS \citep{WS-InterpretableML-NIPS16,WS-InterpretableML-NIPS17,WS-InterpretableML2-NIPS17} or CHI \citep{WS-HumanCenteredLearning-CHI16}. Moreover, several books on the subject have already appeared or an in preparation
\citep{ExplainableInterpretableModels,InterpretableMachineLearning}, funding agencies like DARPA have
recognized the need for explainable
AI\footnote{\url{http://www.darpa.mil/program/explainable-artificial-intelligence}}, 
and the General Data Protection Regulation of the EC includes a ''right to explanation'', which may have a strong impact on machine learning and data mining solutions \citep{GPDR-Piatetsky}.

The strength of many recent
learning algorithms, most notably deep learning
\citep{DeepLearning-Nature,DeepLearning-Overview}, feature learning
\citep{Word2Vec}, fuzzy systems \citep{Interpretability-FuzzySystems} or topic modeling \citep{ProbabilisticTopicModels},
is that latent variables are formed during the learning
process. Understanding the meaning of these hidden variables is
crucial for transparent and justifiable decisions.  Consequently,
visualization of such model components has recently received some
attention \citep{VisualizingTopicModels,Zeiler2014,rothe2016}.
Alternatively, some research has been devoted to trying to convert
such arcane models to more interpretable rule-based 
or tree-based theories
\citep{NN-RuleExtraction,NN-DataMining,ANN-DT,zilke2016deepRED}
or to develop hybrid models that combine the interpretability of logic with the predictive strength of statistical and probabilistic models
\citep{LogicUncertainty,DeepLogicNetworks,HarnessingDNNs}.

Instead of making the entire model interpretable, methods like LIME
\citep{LIME} are able to provide local explanations for
inscrutable models, allowing to trade off fidelity to the original
model with interpretability and complexity of the local model.
In fact, \citet{TextClassification-Explanations} report on experiments that
illustrate that such local, instance-level explanation are preferable to global, document-level models. An interesting aspect of rule-based theories is that they
can be considered as hybrids between
local and global explanations \citep{jf:Dagstuhl-04}: A rule set may be viewed as a global model, whereas the individual rule that fires for a particular example may be viewed as a local explanation.

\enlargethispage*{12pt}

Nevertheless, in our view, many of these approaches fall short in that they take the interpretability of rule-based models for granted.
Interpretability is often considered to
correlate with complexity, with the intuition that simpler models are
easier to understand.    
Principles like Occam's Razor
\cite{OccamsRazor} or Minimum Description Length (MDL) \cite{MDL}
are commonly used heuristics for model selection, and have shown to be
successful in overfitting avoidance. As a consequence, most rule
learning algorithms have a strong bias towards simple theories.
Despite the necessity of a bias for simplicity
for overfitting avoidance, we argue in this paper that simpler rules are not
necessarily more interpretable,
at least not when other aspects of interpretability beyond the mere syntactic readability are considered.
This implicit equation of comprehensibility and simplicity was already criticized by, e.g.,  \citet{850821}, who argued that ''\emph{there has been no study that shows that people find
smaller models more comprehensible or that the size of a model is the
only factor that affects its comprehensibility.}''
There are also a few systems that explicitly strive
for longer rules, and recent evidence has shed some doubt on the
assumption that shorter rules are indeed
preferred by human experts. We will discuss the relation of
rule complexity and interpretability at length in Section~\ref{sec:Interpretability}.

Other criteria than accuracy and model complexity have rarely been
considered in the learning process. For example,
\citet{DM-NLP-14-CoherentRules} proposed to consider the semantic
coherence of its conditions when formulating a rule.
\citet{Rules-MedicalExperts} show that rules that respect monotonicity
constraints are more acceptable to experts than rules that do not. As
a consequence, they modify a rule learner to respect such constraints
by ignoring attribute values that generally correlate well with other
classes than the predicted class. 
\citet{ComprehensibleModels} reviews these and other approaches, compares several classifier types with
respect to their comprehensibility, and points out several drawbacks
of model size as a single measure of interpretability.

In his pioneering 
framework for
inductive learning, 
\citet{InductiveLearning} stressed its links with cognitive science,
noting that 
``\emph{inductive learning has a strong cognitive science flavor}'', and 
postulates that 
``\emph{descriptions generated by inductive inference bear similarity
to human knowledge representations}'' with reference to \citet{hintzman1978psychology}, an elementary
text from psychology on human
learning. \citet{InductiveLearning} considers adherence to the
comprehensibility postulate to be ''\emph{crucial}'' for inductive rule learning,
yet, as discussed above, it is rarely ever explicitly addressed beyond equating it with model simplicity. \citet{ExplainableAI-SocialSciences} makes an important first step by providing a comprehensive review of what is known in the social sciences about explanations and discusses these findings in the context of explainable artificial intelligence.

In this paper, we primarily intend to highlight this gap in machine
learning and data mining research.
In particular, we focus on the \emph{plausibility} of rules, which, in our
view, is an important aspect that contributes to interpretability (Section~\ref{sec:Interpretability}). 
In addition
to the comprehensibility of a model, which we interpret in the sense that 
the user can understand the learned model well enough to be able to manually apply it
to new data, and its justifiability, which specifies whether the model is in line with existing knowledge,
we argue that a good model should also be plausible, i.e., be 
convincing and acceptable to the user. For example, as an extreme case, a 
default model that always predicts the majority class, is very interpretable,
but in most cases not very plausible.  We will argue that
different models may have different degrees of plausibility, even if they
have the same discriminative power. Moreover, we believe that the plausibility 
of a model is---all other things being equal---not related or in some cases even positively correlated with the complexity of a model.

To that end, we also report the results
of a crowdsourcing evaluation of learned rules in four domains  (Section~\ref{sec:crowd}). Overall, the performed experiments are based on nearly 3,000 judgments collected from 390 distinct participants.
The results show that there is indeed no evidence that shorter rules are preferred by humans. On the contrary, we could observe a preference for longer rules in two of the studied domains (Section~\ref{sec:complexity}). In the following, we then relate this finding to 
related results in the psychological literature, such as the conjunctive fallacy (Section~\ref{sec:conjunction-fallacy}) and insensitivity to sample size (Section~\ref{sec:insensitivity}). Section~\ref{sec:relevance} is devoted to a discussion of the relevance of conditions in rules, which may not always have the expected influence on one's preference, in accordance with the recently described weak evidence effect. The remaining sections focus on the interplay of cognitive factors and machine readable semantics: Section~\ref{sec:recognition} covers the recognition heuristic, 
Section~\ref{sec:coherence} discusses the effect of semantic coherence on interpretability, and Section~\ref{sec:structure} briefly highlights the lack of methods for learning structured rule-based models.

\section{Aspects of Interpretability}
\label{sec:Interpretability}


\label{sec:interpretability}
Interpretability is a very elusive concept which we use
in an intuitive sense. 
\citet{ComprehensibilityManifesto} has already
observed that it is an ill-defined concept, and has called
upon several communities from both academia and industry to tackle this problem, to ''find objective definitions of what comprehensibility is'', and to open ''the hunt for probably approximate comprehensible learning''.
Since then, not much has changed.
For example, \citet{lipton2016mythos} still suggests that the term interpretability is ill-defined. In fact,
the concept can be
found under different names in the literature, including
understandability, interpretability, comprehensibility, plausibility, 
trustworthiness, justifiability and others. They all have slightly different
semantic connotations. 

A thorough clarification of this terminology is beyond the 
scope of this paper, but in the following, we briefly highlight different
aspects of interpretability, and then proceed to clearly define and distinguish
comprehensibility and plausibility, the two aspects that are pertinent to this work.

\subsection{Three Aspects of Interpretability}
\label{sec:dimensions}

\begin{figure}[t]
	\centering
	\resizebox{0.9\textwidth}{!}{
		\includegraphics{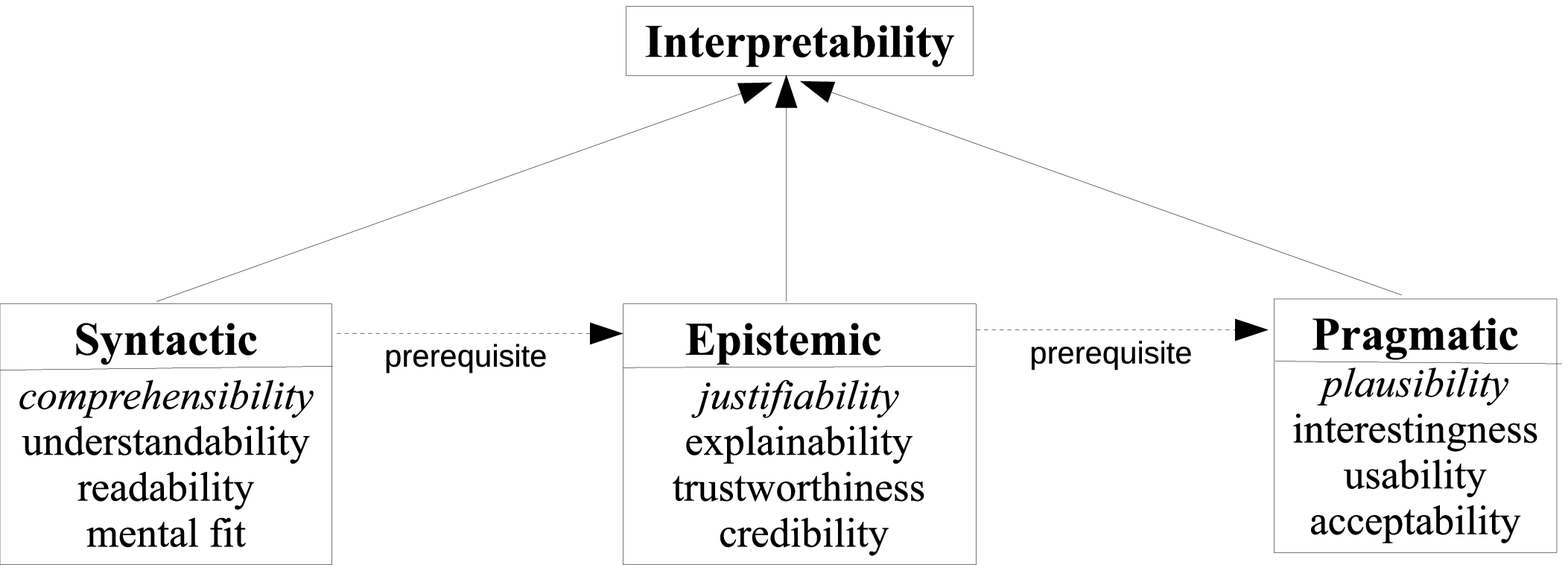}}
	\caption{Three aspects of interpretability}
	\label{fig:interpretability-aspects}
\end{figure}

In this section, we attempt to bring some order into the 
multitude of terms that are used in the context of interpretability. Essentially, we distinguish three aspects of interpretability (see also  Figure~\ref{fig:interpretability-aspects}):
\begin{description}
	\item[\it syntactic interpretability:] This aspect is concerned with the ability of the user to 
comprehend the knowledge that is encoded in the model, in very much the same way
as the definition of a term can be understood in a conversation or a textbook. 
   \item[\it epistemic interpretability:] This aspect assesses
    to what extent the model is in line with existing domain knowledge. A model can be interpretable in the sense that the user can operationalize and apply it, but
    the encoded knowledge or relationships are not well correlated with the user's prior knowledge. For example, a model which states that the temperature is rising on odd-numbered days and falling on even-numbered days has a high syntactic interpretability but a low epistemic interpretability.	
   \item[\it pragmatic interpretability:] Finally, we argue that it is important to 
   capture whether the model serves the intended purpose. 
   A model can be perfectly interpretable in the syntactic and epistemic sense, but have a low pragmatic value for the user. 
   For example, the simple model that the temperature tomorrow will be roughly the same as
   today is obviously very interpretable in the syntactic sense,
   it is also quite consistent with our experience and therefore interpretable
   in the epistemic sense, but 
   it may not be satisfying as an acceptable explanation
   for a weather forecast.
\end{description}

Note that these three categories essentially correspond to the grouping of terms pertinent to interpretability which has previously been introduced by 
\citet{bibal2016interpretability}. 
They treat terms like \emph{comprehensibility}, \emph{understandability}, and \emph{mental fit}, as essentially synonymous to \emph{interpretability}, and use them to denote syntactic interpretability.
In a second group, \citet{bibal2016interpretability} bring notions such as \emph{interestingness}, \emph{usability}, and \emph{acceptability} together, which essentially corresponds to our notion of pragmatic interpretability. Finally, they have \emph{justifiability} as a separate category, which essentially corresponds to what we mean by epistemic interpretability. We also subsume their notion of \emph{explanatory} as
\emph{explainability} in this group, which we view as synonymous to \emph{justifiability}.
A key difference to their work lies in our view that all three of the above are different aspects of interpretability, whereas \citet{bibal2016interpretability}
view the latter two groups as different but related concepts.

We also note in passing that this distinction loosely corresponds to prominent philosophical treatments of explanations \cite{Explanations-Theories}. Classical theories, such as the deductive-nomological theory of explanation \cite{Explanations-Logic}, are based on the validity of the logical connection between premises and conclusion. Instead, \citet{Explanations-Pragmatics} suggests a pragmatic theory of explanations, according to which the explanation should provide the answer to a (why-)question. Therefore, the same proposition may have different explanations, depending on the information demand. For example, an explanation for why a patient was infected with a certain disease may relate to her medical conditions (for the doctor) or to her habits (for the patient).
Thus, pragmatic interpretability is a much more subjective and user-centered notion than epistemic interpretability.

However, clearly these aspects are not independent. As already noted by 
\citet{bibal2016interpretability}, syntactic interpretability is a prerequisite to 
the other two notions. Moreover, we also view epistemic interpretability as a prerequisite to pragmatic interpretability: In case a model is not in line with the user's prior knowledge and therefore has a low epistemic value, it also will have a low pragmatic value to the user. 
Moreover, the differences between the terms shown in Figure~\ref{fig:interpretability-aspects} are soft,
and not all previous studies have used them in consistent ways.
For example, \citet{muggleton2018ultra} employ a primarily syntactic notion of comprehensibility (as we will see in Section~\ref{sec:comprehensibility}), 
and evaluate it by testing whether the participants in their study can successfully apply the
acquired knowledge to new problems. In addition, it is also
measured whether they can give meaningful names to the explanations they deal with, and
whether these names are helpful in applying the knowledge. Thus, these experiments try to
capture epistemic aspects as well.


\subsection{Comprehensibility}
\label{sec:comprehensibility}

One of the few attempts for an operational definition of interpretability is given in the works of \citet{PI-Comprehensibility} and \citet{muggleton2018ultra}, who related the concept to objective measurements such as the time needed for inspecting a learned concept, for applying it in practice, or for giving it a meaningful and correct name. 
This gives interpretability a clearly syntactic interpretation in the sense defined in Section~\ref{sec:dimensions}.
Following \citet{muggleton2018ultra}, we refer to this type of syntactic interpretability as \emph{comprehensibility}, and define it as follows:

\begin{definition}[Comprehensibility]\label{def:comprehensibility}
	A model $\mdl_1$ is more ''comprehensible'' than a model $\mdl_2$ with respect to a given task if a human user makes fewer mistakes in the application of
	model $\mdl_1$ to new samples drawn randomly from the task domain than
	when applying $\mdl_2$.
\end{definition}
Thus, a model is considered to be comprehensible if a user is able to understand all the mental calculations that are prescribed by the model, and can successfully apply the model to new tasks drawn from the same population. A model is more comprehensible than another model if the user's error rate in doing so is smaller.\footnote{We are grateful to one of our reviewers for pointing out that this essentially is in line with the cognitive science perspective on comprehension as proposed by \citet{NLP-MentalModels}, where understanding a natural language sentence or text means to be able to draw valid conclusions and inferences from it.}
\citet{muggleton2018ultra} study various related, measurable quantities, such 
as the inspection time, the rate with which the meaning of the predicate is
recognized from its definition, or the time used for coming up with a suitable name for a definition.

\paragraph{Relation to Alternative Notions of Interpretability.}
\citet{DecisionTreeComprehensibility} use a very similar definition when they study
how the response time for various data- and model-related tasks such as ''classify'', ''explain'', ''validate'', or ''discover'' varies with changes in the
structure of learned decision trees.
Another variant of this definition was suggested by
\citeauthor{TypifyingIntepretability} (\citeyear{TypifyingIntepretability,InterpretabilityFramework}), who consider
interpretability relative to a target model, typically (but not necessarily)  a
human user. More precisely, they define a learned model as $\delta$-interpretable relative to a target model if the target model can be improved by a factor of $\delta$ (e.g., w.r.t. predictive accuracy) with information obtained by the learned model. 
All these notions have in common that they relate interpretability to a performance aspect, in the sense that
a task can be performed better or performed at all with the help of the learned model. 

As illustrated in Figure~\ref{fig:interpretability-aspects},
we consider understandability, readability and mental fit as alternative terms for syntactic interpretability.
Understandability is considered as a direct synonym for comprehensibility \cite{bibal2016interpretability}. Readability clearly corresponds to syntactic level.
The term  \emph{mental fit} may require additional explanation. We used it in the sense of \citet{RuleInduction-IDA} to denote suitability of the representation (i.e. rules) for a given purpose (to explain a classification model).

\subsection{Justifiability}
\label{sec:justifiability}

A key aspect on interpretability is that a concept is consistent with available domain knowledge, which we call epistemic interpretability. 
\citet{justifiability} have introduced this concept under the name of \emph{justifiability}.
They 
consider a model to be more justifiable 
if it better conforms to domain knowledge, which may be viewed as constraints to which a justifiable model has to conform (hard constraints) or should better conform (soft constraints).
\citet{justifiability2} provide a taxonomy of such constraints, which include univariate constraints such as monotonicity as well as multivariate constraints such as preferences for groups of variables.

We paraphrase and slightly generalize this notion in the following definition:

\begin{definition}[Justifiability]
	A model $\mdl_1$ is more ''justifiable'' than a model $\mdl_2$ if $\mdl_1$ violates fewer constraints that are imposed by the user's prior knowledge. 
\end{definition}

%
\citet{justifiability2} also define an objective measure for justifiability, which essentially corresponds to a weighted sum over the fractions of cases where each variable is needed in order to discriminate between different class values.

\paragraph{Relation to Comprehensibility and Plausibility.}

Definition~1 (comprehensibility) addresses the syntactical level of understanding, which is is a prerequisite for justifiability. 
What this definition does not cover are facets of interpretability that relate to one's background knowledge. 
For example, an empty model or a default model, classifying all examples as positive, is very simple to interpret,
comprehend and apply, but such model will hardly be justifiable.

Clearly, one needs to be able to comprehend the definition of a concept before it can be checked whether it corresponds to existing knowledge.
Conversely, we view justifiability as a prerequisite to our notion of plausibility, which we will define more precisely in the next section: a theory that does not conform to domain knowledge is not plausible, but, on the other hand, the user may nevertheless assess different degrees of plausibility to different explanations that are all consistent with our knowledge. In fact, many scientific and in particular philosophical debates are about different, conflicting theories, which are all justifiable but have different degrees of plausibility for different groups of people. 

\paragraph{Relation to Alternative Notions of Interpretability.}
Referring to Figure~\ref{fig:interpretability-aspects}, we view plausibility as an aspect of epistemic interpretability, similar to notions like
explainability, trusthworthiness and credibility. 
Both trustworthiness and credibility imply evaluation of the model against domain knowledge. Explainability is harder to define and has received multiple definitions in the literature. 
We essentially follow \citet{InterpretabilityExplainability}, who makes a distinction that is similar to our notions of syntactic and epistemic interpretability: in his view, interpretability is to allow the user to grasp the mechanics of a process (similar to the notion of mental fit that we have used above), whereas explainability also implies a deeper understanding of why the process works in this way. This requires the ability to relate the notion to existing knowledge, which is why we view it primarily as an aspect of epistemic interpretability.


\begin{figure}[t]
	\begin{Verbatim}
	QOL = High :- Many events take place. 
	QOL = High :- Host City of Olympic Summer Games. 
	QOL = Low  :- African Capital.
	\end{Verbatim}
	\centering (a) rated highly by users
	
	\begin{Verbatim}
	QOL = High :- # Records Made >= 1, # Companies/Organisations >= 22.
	QOL = High :- # Bands >= 18, # Airlines founded in 2000 > 1.
	QOL = Low  :- # Records Made = 0, Average January Temp <= 16.
	\end{Verbatim}
	
	\centering (b) rated lowly by users
	
	\caption{Good discriminative rules for the quality of living of a city \citep{PaulheimESWC2012b}}
	\label{fig:QOL-rules}
\end{figure}

\subsection{Plausibility}
\label{sec:plausibility}

In this paper, we focus on a pragmatic aspect of interpretability,
which we refer to as \emph{plausibility}. We primarily view this notion in the  sense of ''user acceptance'' or ''user preference''. However, as discussed in Section~\ref{sec:dimensions}, this also means that it has to rely on aspects of syntactic and epistemic interpretability as prerequisites.
%
 For the purposes of this paper, we 
 define plausibility as follows:

\begin{definition}[Plausibility]
A model $\mdl_1$ is more ''plausible'' than a model $\mdl_2$ if $\mdl_1$ is more likely to be accepted by a user than $\mdl_2$.
\end{definition}
Within this definition, the word ``accepted'' bears the meaning specified by the Cambridge English Dictionary\footnote{\url{https://dictionary.cambridge.org/dictionary/english/accepted}} as
``generally agreed to be satisfactory or right''.

Our definition of plausibility is less objective than the above definition of
comprehensibility because it always relates to the subject's perception of the
utility of a given explanation, 
i.e., its pragmatic aspect.
Plausibility, in our view, is inherently subjective, i.e., it relates to 
the question how useful a model is perceived by a user. Thus, it
needs to be evaluated in introspective user studies, where the
users explicitly indicate how plausible an explanation is, or which of two 
explanations appears to be more plausible. Two explanations that can equally well
be applied in practice (and thus have the same syntactic interpretability) and are both consistent with existing knowledge (and thus have the same epistemic interpretability), may nevertheless be perceived as having different
degrees of plausibility.

\paragraph{Relation to Comprehensibility and Justifiability.}

A model may be consistent with domain knowledge, but nevertheless appear implausible.
Consider, e.g., the rules shown in Figure~\ref{fig:QOL-rules}, which have been derived by the Explain-a-LOD system \citep{paulheim2012unsupervised}. 
The rules provide several possible explanations
for why a city has a high quality of living, using Linked Open Data as background knowledge.
Clearly, all rules are comprehensible
and can be easily applied in practice. They also appear to be justifiable, in the sense that all of them appear to be consistent with prior knowledge. For example, while the number of records made in a city is certainly not a \emph{prima facie} aspect of its quality of living, it is reasonable to assume a correlation between these two variables.
Nevertheless, 
the first three rules appear to be more plausible to a human user, which was
also confirmed in an experimental study \citep{PaulheimESWC2012,PaulheimESWC2012b}. 


\paragraph{Relation to Alternative Notions of Interpretability.}
In Figure~\ref{fig:interpretability-aspects} we consider  \emph{interestingness}, \emph{usability}, and \emph{acceptability} as related terms. All these notions imply some degree of user acceptance  or fitness for given purpose.
%

In the remainder of the paper, we will 
typically talk
about ''plausibility'' in the sense defined above, but we will sometimes
use terms like ''interpretability'' as a somewhat more general term. 
We also use ''comprehensibility'', mostly when we refer to syntactic interpretability, as discussed and defined above.
However, all terms are meant to be interpreted in an intuitive, and non-formal way.\footnote{In particular, we do not intend to touch upon formal notions of plausibility, such as those given in the Dempster-Shafer theory, where plausibility of an evidence is defined as an upper bound on the belief in the evidence, or more precisely, as the converse of one's belief in the opposite of the evidence \cite{DempsterShafer-1,DempsterShafer-2}.}


\section{Setup of Crowdsourcing Experiments on Plausibility}
\label{sec:crowd}

In the remainder of the paper, we focus on the plausibility of rules. In particular, we report on a series of five crowdsourcing experiments, which relate the perceived
plausibility of a rule to various factors such as rule complexity, attribute importance or centrality.
As a basis we used pairs of rules generated by machine learning
systems, typically one rule representing a shorter, and the other a longer explanation.
Participants were then asked to indicate which one of
the pair they preferred. 

The selection of crowdsourcing as a means of acquiring data allows us
to gather thousands of responses in a manageable time frame while at
the same time ensuring our results can be easily replicated.\footnote{To this end, source datasets, preprocessing code, the responses obtained with crowdsourcing, and the code used to analyze them were made available at \url{https://github.com/kliegr/rule-length-project}. The published data do not contain quiz failure rates (\emph{qfr} in Table~\ref{tbl:cf-rulelength-statistics}--Table~\ref{tab:pagerank}), since these were computed from statistics only displayed in the dashboard of the used crowdsourcing platform 
	upon completion of the crowdsourcing tasks.}
In the following, we describe the basic setup that is common to all performed experiments.
Most of the setup is shared for the subsequent experiments and will not be repeated, 
only specific deviations will be mentioned.
Cognitive science research has different norms for describing experiments than those that are commonly employed in machine learning research.\footnote{In fact, with \emph{psychometrics}, an entire field is devoted to proper measurement of psychological phenomena \cite{Psychometrics-Introduction}.}
Also, the parameters of the experiments, such as the amount of payment, is described in somewhat greater detail than usual in machine learning, because of the general sensitivity of the participants to such conditions.

We tried to respect these differences by dividing experiment descriptions here and in subsequent sections into subsections entitled \emph{''Material''}, \emph{''Participants''}, \emph{''Methodology''}, and \emph{''Results''}, which correspond to the standard outline of an experimental account in cognitive science. In the following, we describe the general setup that applies to all experiments in the following sections, where then the main focus can be put on the results.  

\subsection{Material}
\label{sec:material}

For each experiment, we generated rule pairs generated with two different learning algorithms, and asked users about their preference. The details of the rule generation and selection process are described in this section.

\subsubsection{Domains}
\label{sec:domains}

\enlargethispage*{12pt}

For the experiment, we used learned rules in four domains (Table~\ref{tbl:cf-dataset-overview}):

\begin{table}[t]
	\begin{center}
		\caption{Overview of the datasets used for generating rule pairs}
		\label{tbl:cf-dataset-overview}
		\medskip
		\begin{tabular}{cllrrl}
			\hline\noalign{\smallskip}
			\# pairs & dataset & data source & \# instances & \# attr. & target \\
			\noalign{\smallskip}\hline\noalign{\smallskip}
			80 & \data{Traffic} & LOD & 146 & 210 & rate of traffic accidents in a country\\
			36 & \data{Quality} & LOD & 230 & 679  & quality of living  in a city \\
			32 & \data{Movies} & LOD & 2000 & 1770  & movie rating \\
			10 & \data{Mushroom} & UCI & 8124 &23 & mushroom poisonous/edible \\
			\noalign{\smallskip}\hline
		\end{tabular}
	\end{center}
\end{table}

\begin{description}
	\item[\data{Mushroom}] contains mushroom records
	drawn from Field Guide to North American Mushrooms
	\citep{lincoff1989audubon}. Being available at the UCI repository \cite{UCI}, it is arguably one of the most frequently used
	datasets in rule learning research, its main advantage being discrete,
	understandable attributes. 
	\item[\data{Traffic}] is a statistical dataset of death rates in traffic accidents by country, obtained from the WHO.\footnote{\url{http://www.who.int/violence_injury_prevention/road_traffic/en/}}
	\item[\data{Quality}] is a dataset derived from the Mercer Quality of Living index, which collects the perceived quality of living in cities world wide.\footnote{\url{http://across.co.nz/qualityofliving.htm}}
	\item[\data{Movies}] is a dataset of movie ratings obtained from MetaCritic.\footnote{\url{http://www.metacritic.com/movie}}
\end{description}
The last three datasets were derived from the Linked
Open Data (LOD) cloud \cite{SemanticWeb-Datasets}. 
Originally, they
consisted only of a name and a target variable, such as a city and its quality-of-living index, or a movie and its rating. The names were then linked to entities in the public LOD dataset DBpedia, using the method described by \citet{paulheim2012unsupervised}. From that dataset, we extracted the classes to which the entities belong, using the deep classification of YAGO, which defines a very fine grained class hierarchy of several thousand classes. Each class was added as a binary attribute. For example, the entity for the city of Vienna would get the binary features \emph{European Capitals}, \emph{UNESCO World Heritage Sites}, etc.

The goal behind these selections was that the domains are general enough so that the participants are able to comprehend a given rule
without the need for additional background knowledge, but are nevertheless not able to reliably judge
the validity of a given rule. Thus, participants will need to rely on their common sense
in order to judge which of two rules appears to be more convincing.
This also implies that we specifically did not expect the users to have expert knowledge in
these domains. 

\subsubsection{Rule Generation}
\label{sec:rule-generation}

We used
two different approaches to generate rules for each of the four domains mentioned in the previous section.
\begin{description}
	\item[\it Class Association Rules:] We used a standard implementation of
	the \alg{Apriori} algorithm for association rule learning
	\citep{DBLP:conf/sigmod/AgrawalIS93,hahsler2011arules} and filtered
	the output for class association rules with a minimum support of $0.01$, minimum confidence of $0.5$, and a maximum length of $5$. Pairs were formed between all rules that correctly classified at least one shared instance.
	Although other more sophisticated approaches (such as a threshold on the
	Dice coefficient) were considered, it turned out that the process
	outlined above produced rule pairs with quite similar values of
	confidence (i.e. most equal to $1.0$), except for the \data{Movies} dataset.
	\item[\it Classification Rules:] We used a simple top-down greedy
	hill-climbing algorithm
	that takes a seed example and generates a pair of rules, one with a
	regular heuristic (Laplace) and one with its inverted
	counter-part. As shown by \citet{jf:DS-16-ShortRules} and
	illustrated in Figure~\ref{fig:mushroom}, this
	results in rule pairs that have approximately the same degree of
	generality but different complexities.
\end{description}


From the resulting rule sets, we selected several \emph{rule pairs} consisting of a long and
a short rule that have the same or a similar degree of generality.\footnote{The generality of a rule is defined via the set of examples a rule covers. Two rules that cover the same \emph{set} of examples have the same  generality, even if they have a different number of conditions. Examples include an itemset and its closure, or the elephant example discussed further below in Section~\ref{sec:bias-complexity}. We use the phrase ''degree of generality'' somewhat loosely to refer to two rules that cover an equal \emph{number} of examples, such 
	as the pair of first rules of the two solutions in the \data{Mushroom} dataset (Figure~\ref{fig:mushroom}).} 
For \data{Quality} and \data{Movies}, all rule pairs were used. For
the \data{Mushroom} dataset, we selected rule pairs so that every
difference in length (one to five) is represented.
All selected rule pairs were pooled, so we did not discriminate between the 
learning algorithm that was used for generating them.
For the \data{Traffic} dataset the rule learner generated a higher number of rules than for the other datasets, which allowed us to select the rule pairs for annotation
in such a way that various types of differences between rules in each pair were represented. Since this stratification procedure, detailed in \cite{kliegr2017thesis}, applied only to one of the datasets, we do not expect this design choice to have profound impact on the overall results and omit a detailed description here.

\begin{figure}[t]
	\centering
	\resizebox{\textwidth}{!}{
		\includegraphics{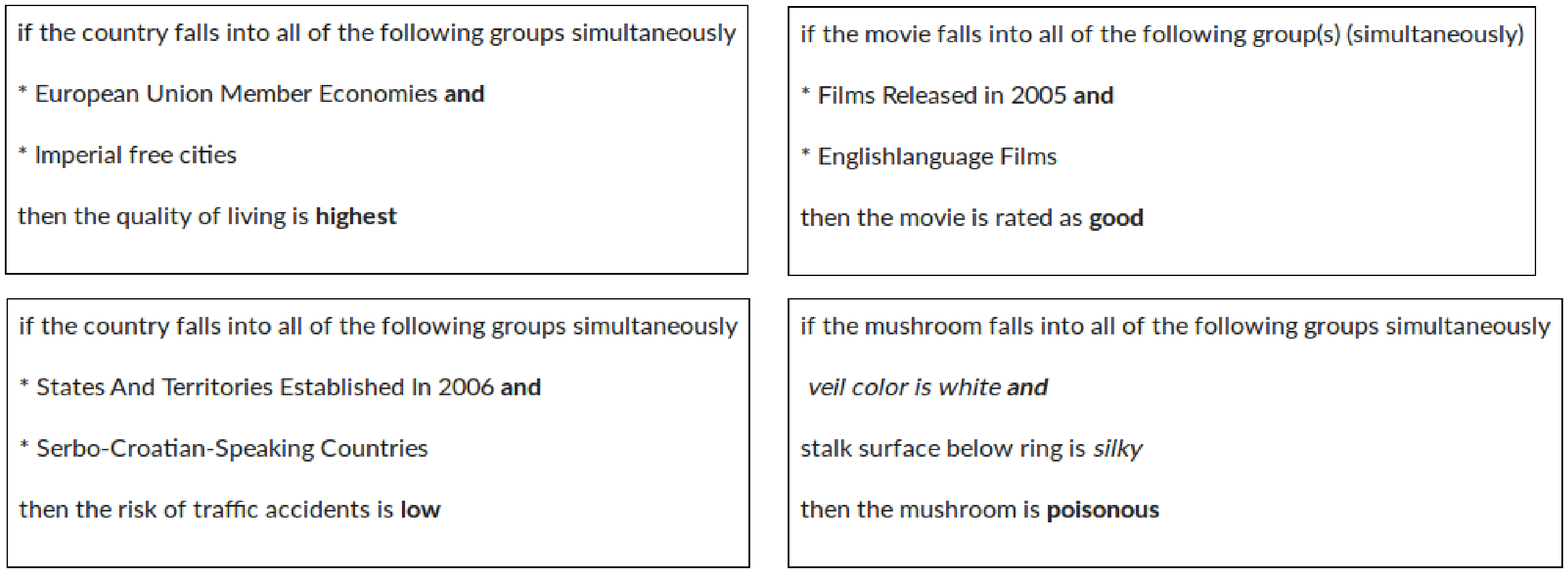}}
	\caption{Example translated rules for the four datasets}
	\label{figure:translations}
\end{figure}

As a final step, we automatically translated all rule pairs into
human-friendly HTML-formatted text, and randomized
the order of the rules in the rule pair. 
Example rules for the four datasets are shown in 
Figure~\ref{figure:translations}.
The first column of Table~\ref{tbl:cf-dataset-overview} shows the final
number of rule pairs generated in each domain.

\subsection{Methodology}
\label{sec:cf:expsetup}

The generated rule pairs were then evaluated in a user study on a crowdsourcing platform, where participants were asked to issue a preference between the plausibility of the shown rules. This was then correlated to various factors that could have an influence on plausibility.

\subsubsection{Definition of Crowdsourcing Experiments}
As the experimental platform we used the CrowdFlower crowdsourcing
service.\footnote{Since our experiments, CrowdFlower has been re-branded under the name \emph{Figure Eight} and is now available at \url{https://www.figure-eight.com/}.} Similar to the
better-known Amazon Mechanical Turk, CrowdFlower allows to distribute
questionnaires to participants
around the world, who complete them for remuneration. The remuneration is typically a small payment in US dollars---for one judgment relating to one rule we paid 0.07 USD---but some participants may receive the payment in other currencies, including in game currencies (``coins'').

\enlargethispage*{12pt}

A crowdsourcing task performed in CrowdFlower consists of a sequence of 
steps: 
\begin{enumerate}
	\item The CrowdFlower platform recruits participants, so-called \emph{workers} for the task from a pool of its users, who match the level and geographic requirements set by the experimenter. The workers decide to participate in the task based on the payment offered and the description of the task.
	\item Participants are presented assignments which contain an illustrative example.
	\item If the task contains test questions, each worker has to pass a \emph{quiz mode} with test questions. Participants learn about the correct answer after they pass the quiz mode, and have the option to contest the correct answer if they consider it incorrect.
	\item Participants proceed to the \emph{work mode}, where they complete the task they have been assigned by the experimenter. The task typically has the form of a questionnaire. If test questions were defined by the experimenter, the CrowdFlower platform randomly inserts test questions into the questionnaire. Failing a predefined proportion of hidden test questions results in removal of the worker from the task. Failing the initial quiz or failing a task can also reduce participants' accuracy on the CrowdFlower platform. Based on the average accuracy, participants can reach one of the three levels. A higher level gives a user access to additional, possibly better paying tasks.
	
	\item Participants can leave the experiment at any time. To obtain payment for their work, they need to submit at least one page of work. After completing each page of work, the worker can opt to start another page. The maximum number of pages per participant is set by the experimenter. As a consequence, two workers can contribute a different number of judgments to the same task.
	\item If a bonus was promised, the qualifying participants receive extra credit. 
\end{enumerate}

\begin{figure}[p!]
	\centering
	\resizebox{\textwidth}{!}{
		\fbox{\includegraphics{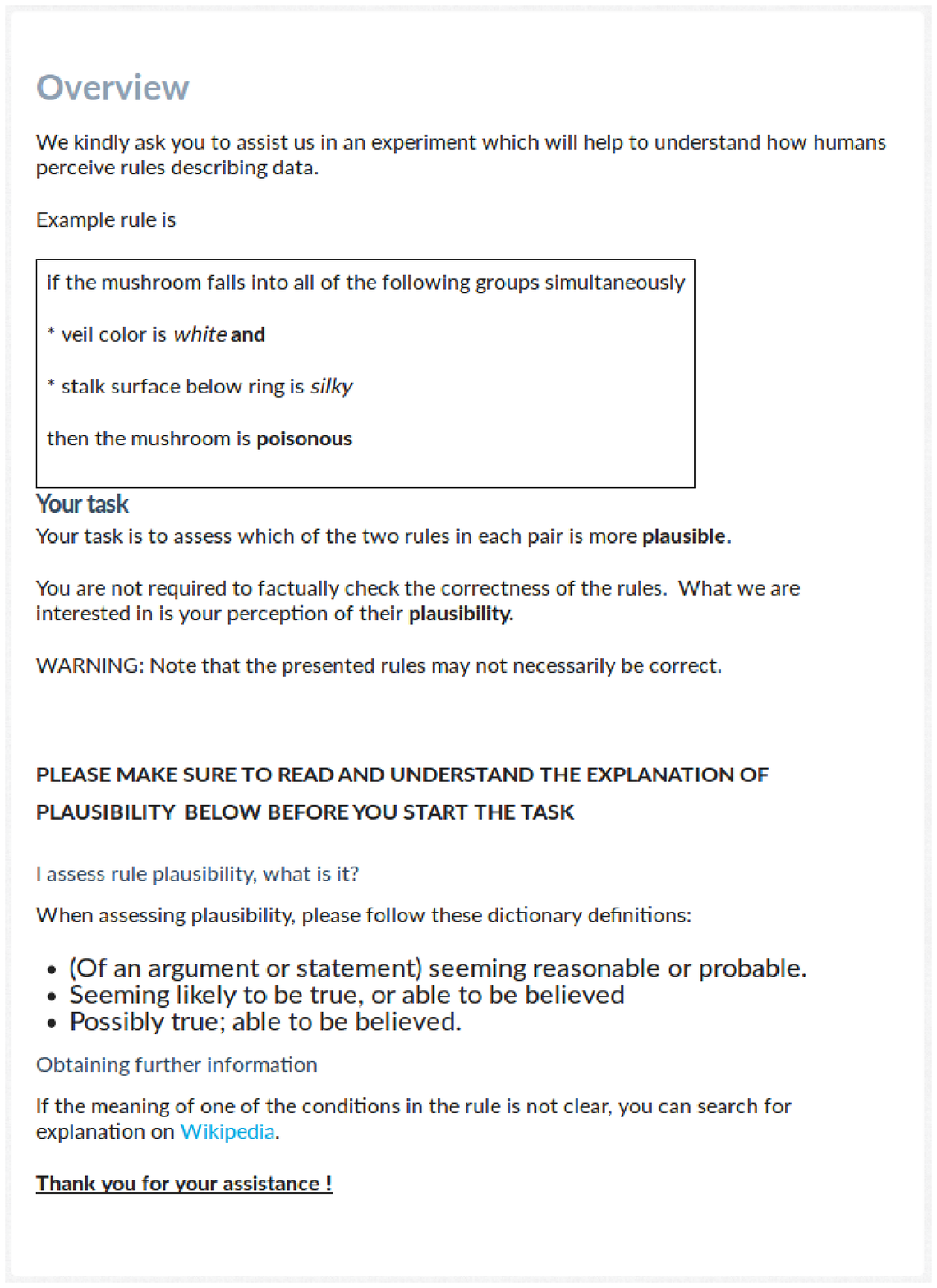}}}
	\caption{Example instructions for experiments 1--3. The example rule pair was adjusted based on the dataset. For Experiment 3, the box with the example rule additionally contained values of confidence and support, formatted as shown in Figure~\ref{fig:supp-conf}.}
	\label{fig:InstructionsComplete}
\end{figure}

The workers were briefed with task instructions, which described the purpose of the task, gave an example rule, and explained plausibility as the elicited quantity (cf.\ Figure~\ref{fig:InstructionsComplete}). As part of the explanation, the participants were given definitions of ``plausible'' sourced from the Oxford Dictionary\footnote{\url{https://en.oxforddictionaries.com/definition/plausible}} and the  Cambridge Dictionary\footnote{\url{https://dictionary.cambridge.org/dictionary/english/plausible}} (British and American English). 
The individual task descriptions differed for the five tasks, and will be described 
in more detail later in the paper in the corresponding sections.
Table~\ref{tbl:datatypes} shows a brief overview of the factors variables and their data types for the five experiments.


\begin{table}[tbp]
	\caption{Variables used in Experiment 1--5. $\Delta$ after a variable refers to the difference 
		of its values of a given rule pair.
		Plausibility was elicited on a five-level linguistic scale
		ranging from $-2$ for ``Rule~2 (strong preference)'' to $+2$ for ``Rule~1 (strong preference)''.}
	\begin{tabular}{rp{6cm}lll}
		\toprule
		\multicolumn{1}{l}{} &\multicolumn{2}{c}{independent variable}    & \multicolumn{2}{c}{dependent variable} \\ 
		\multicolumn{1}{l}{Exp.} & name & data type & name&data type \\ 
		\midrule
		1 & rule length $\Delta$ & continuous & plausibility&discrete \\ 
		2 & rule length $\Delta$ & continuous & plausibility&discrete \\ 
		3 & rule support $\Delta$, rule confidence $\Delta$ & discrete & plausibility&discrete \\ 
		4 & attribute importance avg $\Delta$, att. imp. max $\Delta$, literal imp. avg $\Delta$, lit. imp. max $\Delta$ & continuous & plausibility &discrete \\ 
		5 & literal PageRank avg $\Delta$,  literal PageRank max $\Delta$ & continuous & plausibility&discrete \\ 
		\bottomrule
	\end{tabular}
	\label{tbl:datatypes}
\end{table}

\subsubsection{Evaluation}
\enlargethispage*{12pt}
Rule plausibility was elicited on a five-level linguistic scale
ranging from ``Rule~2 (strong preference)'' to ``Rule~1 (strong preference)'',
which were interpreted as ordinal values from $-2$ to $+2$.
Evaluations were performed at the level of individual judgments, also called micro-level,  i.e., each response was considered to be a single data point, and multiple judgments for the same pair were not aggregated prior to the analysis.
By performing the analysis at the micro-level, we avoided the possible loss of information as well as the aggregation bias \citep{clark1976effects}. Also, as shown for example by \citet{robinson-ecological}, the ecological (macro-level) correlations are generally larger than the micro-level correlations, therefore by performing the analysis on the individual level we obtain more conservative results. 

We report 
rank correlation between a factor and the observed evaluation (Kendall's $\tau$, Spearman's $\rho$)
and tested whether the coefficients are significantly different from zero.  We will refer to the values of Kendall's $\tau$ as the primary measure of rank correlation, since according to \citet{gibbons1990rank} and \citet{newson2002parameters}, the confidence intervals for Spearman's $\rho$  are less reliable than confidence intervals for Kendall's $\tau$. 

For all obtained correlation coefficients we compute the $p$-value, which is the probability of obtaining a correlation coefficient at least as extreme as the one that was actually observed assuming that the null hypothesis holds, i.e., that there is no correlation between the two variables.  The typical cutoff value for rejecting the null hypothesis is $\alpha=0.05$.




\subsection{Participants}

\enlargethispage*{12pt}

The workers in the CrowdFlower platform were invited to participate in individual tasks.
CrowdFlower divides the available workforce into three levels depending on the accuracy they obtained on earlier tasks. 
As the level of the CrowdFlower workers we chose Level~2, which was described as follows: ``\emph{Contributors in Level~2 have completed over a hundred Test Questions across a large set of Job types, and have an extremely high overall Accuracy.}''.

In order to avoid spurious answers, we also employed a minimum threshold of 180 seconds for completing a page; workers taking 
less than this amount of time to complete a page
were removed from the job. A maximum time required to complete the assignment was not specified, and 
the maximum number of judgments per contributor was not limited.

For quality assurance, each participant who decided to accept the task
first faced a quiz consisting of a random selection of
previously defined 
test questions.
These
had the same structure as regular questions but additionally contained
the expected correct answer (or answers) as well as an explanation for
the answer. We used \emph{swap test} questions where the order of the
conditions was randomly permuted in each of the two pairs, so that 
the participant should not have a preference for
either of the two versions.
The correct answer and explanation was only shown after the worker had responded to the question. 
Only workers achieving at least 70\% accuracy on test questions could proceed to the main task. 

\begin{table}[t]
	\caption{Geographical distribution of collected judgments}\label{tbl:workers}
	\centering
	\medskip
	\subfloat[Experiments 1--3]{%
		\begin{tabular}{lrrrrrrlllc}
			\toprule
			& \multicolumn{3}{c}{Group 1 judgments} & \multicolumn{3}{c}{Group 2 judgments}  &  \multicolumn{3}{c}{Group 3 judgments} &   \multicolumn{1}{c}{total} \\ 
			& USA & UK & Can & USA & UK & Can & USA & UK & Can & judgments \\ 
			\midrule
			\data{Quality} & 68 & 64 & 52 & 96 & 40 & 44 &  &  &  & 364 \\ 
			\data{Movies} & 80 & 52 & 28 & 76 & 30 & 58 & 84 & 44 & 32 & 484 \\ 
			\data{Traffic} & 204 & 120 & 84 & 212 & 116 & 72 &  &  &  & 808 \\ 
			\data{Mushroom} & 106 & 84 & 60 & 97 & 21 & 32 &  &  &  & 400 \\ 
			\midrule
			total & 458 & 320 & 224 & 481 & 207 & 206 & 84 & 44 & 32 & $\mathbf{2056}$ \\ 
			\bottomrule
		\end{tabular}
		\label{tbl:exp1-3workers}
	}
	
	%
	\subfloat[Experiments 4--5]{
		\centering
		\medskip
		\begin{tabular}{lrrrrrrc}
			\toprule
			& \multicolumn{3}{c}{Literal relevance}  & \multicolumn{3}{c}{Attribute relevance}  & \multicolumn{1}{c}{total} \\ 
			& \multicolumn{1}{l}{USA} & \multicolumn{1}{l}{UK} & \multicolumn{1}{l}{Can} & USA & UK & Can & \multicolumn{1}{c}{judgments} \\ 
			\midrule
			\data{Quality} & 63 & 65 & 37 &  &  &  & 165 \\ 
			\data{Movies} & 74 & 46 & 30 &  &  &  & 150 \\ 
			\data{Traffic} & 164 & 58 & 68 & 0 & 10 & 25 & 325 \\ 
			\data{Mushroom} & 70 & 56 & 44 & 23 & 31 & 38 & 262 \\ 
			\midrule
			total & 371 & 225 & 179 & 23 & 41 & 63 & $\mathbf{902}$ \\ 
			\bottomrule
		\end{tabular}
		\label{tbl:exp4-5workers}
	}
\end{table}

\subsubsection{Statistical Information about Participants}
\label{ss:cf:statistical}
CrowdFlower does not publish demographic data about its base of workers. Nevertheless, for all executed tasks, the platform makes available the location of the worker submitting each judgment.  In this section, we use this data to elaborate on the number and geographical distribution of workers participating in Experiments 1--5 described later in this paper.

Table~\ref{tbl:exp1-3workers} reports on workers participating in Experiments 1--3, where  three types of guidelines were used in conjunction with four different datasets, resulting in 9 tasks in total (not all combinations were tried).  
Experiments 4--5 involved 
different guidelines (for determining attribute and literal relevance) and the same datasets. The geographical distribution is reported in Table~\ref{tbl:exp4-5workers}. In total, the reported results are based on 2958 \emph{trusted} judgments.\footnote{A trusted judgment is an answer from a worker that passed the initial quiz and on submitting the work had accuracy score higher than preset minimum accuracy on any hidden test questions. Only trusted judgments were used for analyses.}
Actually, more judgments were collected, but some were excluded due to automated quality checks. 

In order to reduce possible effects of language proficiency, we restricted our participants to English-speaking countries.
Most judgments (1417) were made by workers from United States, followed by the United Kingdom (837) and Canada (704).
The number of distinct participants for each crowdsourcing task is reported in detailed tables describing the results of the corresponding experiments (\emph{part} column in Tables~\ref{tbl:cf-rulelength-statistics}--\ref{tab:pagerank}). Note that some workers participated in multiple tasks. The total number of distinct participants across all tasks reported in Tables~\ref{tbl:exp1-3workers} and~\ref{tbl:exp4-5workers} is 390.

\subsubsection{Representativeness of Crowdsourcing Experiments}
\label{ss:cf:worker}
There is a number of differences between crowdsourcing and the controlled laboratory environment previously used to run psychological experiments. The central question is to what extent do the cognitive abilities and motivation of participants differ between the crowdsourcing cohort and the controlled laboratory environment.
Since there is a small amount of research specifically focusing on the population of the CrowdFlower platform, which we use in our research, we present data related to Amazon Mechanical Turk, under the assumption that the descriptions of the populations will not differ substantially.\footnote{This is supported by the fact that until about 2014, CrowdFlower platform involved Amazon Mechanical Turk (AMT) workers. As of 2017, these workers are no longer involved, because according to CrowdFlower, the AMT channel was both slower and less accurate than other channels used by the CrowdFlower platform (cf. \url{http://turkrequesters.blogspot.com/2014/01/crowdflower-dropping-mechanical-turk.html}).}
This is also supported by previous work such as \cite{MechanicalTurk-CrowdFlower}, which has indicated that the user distribution of CrowdFlower and AMT is comparable.

The population of crowdsourcing workers is a subset of the population of Internet users, which is described in a recent meta study by \citet{paolacci2014inside} as follows: \emph{``Workers tend to be younger  (about 30 years old), overeducated, underemployed, less religious,  and  more  liberal  than  the  general  population.''} While there is limited research on workers' cognitive abilities, \citet{paolacci2010running} found ``\emph{no difference between workers,  undergraduates, and other Internet users on a self-report  measure  of  numeracy  that  correlates  highly  with  actual   quantitative abilities.}'' According to a more recent study by \citet{crump2013evaluating}, workers learn more slowly than university students and may have difficulties with complex tasks. 
Possibly the most important observation related to the focus of our study is that according to 
\citet{paolacci2010running} crowdsourcing workers \emph{``exhibit the classic heuristics and biases and pay attention to directions at least as much as subjects from traditional sources.''}

\section{Interpretability, Plausibility, and Model Complexity}
\label{sec:complexity}

The rules shown in Figure~\ref{fig:QOL-rules} may suggest that simpler rules
are more acceptable than longer rules because the highly rated rules (a) are
shorter than the lowly rated rules (b).
In fact, there are many good reasons why simpler models should be
preferred over more complex models.
Obviously, a shorter model can be
interpreted with less effort than a more complex model of the same
kind, in much the same way as reading one paragraph is quicker than
reading one page. Nevertheless, a page of elaborate explanations may
be more comprehensible than a single dense paragraph
that provides the same information (as we all know from reading
research papers).

Other reasons for preferring simpler models include 
that they are easier to falsify,
that there are fewer simpler theories than complex theories, so the a
priori chances that a simple theory fits the data are lower, or that
simpler rules tend to be more general, cover more examples and their
quality estimates are therefore statistically more
reliable. 

However, one can also find results that throw doubt
on this claim. In particular in cases where not only syntactic interpretability 
is considered, there are some previous works where it was observed that
longer rules are preferred by human experts.
In the following, we discuss this issue in some depth,
by first reviewing the use of a simplicity bias in machine
learning (Section~\ref{sec:bias-simplicity}), then taking the alternative point
of view and recapitulating works where more complex theories are
preferred (Section~\ref{sec:bias-complexity}), and then summarizing
the conflicting past evidence for
either of the two views (Section~\ref{ss:comprmodelsize}).
Finally, in Section~\ref{sec:shorter}, 
we report on the results of our first experiment, which aimed at testing
whether rule length has an influence on the interpretability or
plausibility of found rules at all, and, if so, whether people tend to
prefer longer or shorter rules.

\subsection{The Bias for Simplicity}
\label{sec:occam}
\label{sec:bias-simplicity}

\citet{InductiveLearning} already states that inductive learning algorithms need to incorporate a preference criterion for selecting hypotheses  to address the problem of the possibly unlimited number of hypotheses, and that this criterion is typically \emph{simplicity}, referring to philosophical works on simplicity of scientific theories by \citet{kemeny1953use} and \citet{post1960simplicity}, which refine the initial postulate attributed to Ockham, which we discuss further below.
%
According to \citet{post1960simplicity}, judgments of simplicity should not be made ``\emph{solely on the linguistic form of the theory}''.\footnote{\citet{kemeny1953use} gave the example that among competing explanations 	 for the solar system, the model of Tycho Brahe is linguistically simpler than Copernicus' theory because of the convenient choice of the co-ordinate system associated with the heliocentric view.}
This type of simplicity is referred to as \emph{linguistic
  simplicity}. A related notion of \emph{semantic simplicity} is
described through the falsifiability criterion
\citep{popper1935logik,LogicOfScientificDiscovery}, which essentially states that simpler theories can be more easily falsified. Third, \citet{post1960simplicity} introduces \emph{pragmatic simplicity} which relates to the degree to which the hypothesis can be fitted into a wider context.

Machine learning algorithms typically focus on linguistic or syntactic
simplicity, by referring to the description length of the learned
hypotheses.
The complexity of a rule-based model is typically measured with simple
statistics, such as the number of learned rules and their length,
or the total number of conditions in the learned model \citep[cf., e.g.,][]{WRA-Predictive,InterpretableDecisionSets,AntMiner-Evaluation,BayesianRuleSets}.
Inductive rule learning is typically concerned with learning a set of
rules or a rule list which discriminates positive from negative examples
\citep{jf:Book-Nada,RuleML-15}. For this task, a bias towards simplicity is
necessary because for a contradiction-free training set, it is trivial
to find a rule set that perfectly explains the training data, simply
by converting each example to a maximally specific rule that covers
only this example. 

Occam's Razor, \emph{``Entia non
  sunt multiplicanda sine necessitate''},\footnote{Entities should not
  be multiplied beyond necessity.} which is attributed to English
philosopher and theologian William of Ockham (c.\ 1287--1347),
has been put forward as support for a principle of parsimony in the philosophy of science \cite{OccamsRasierer}.
In machine learning, this principle is generally interpreted as
  \emph{``given two
  explanations of the data, all other things being equal, the simpler
  explanation is preferable''} \citep{OccamsRazor}, or simply \emph{``choose the shortest explanation for the observed data''}
  \citep{MachineLearning}.
While it
is well-known that striving for simplicity often yields better
predictive results---mostly because pruning or regularization
techniques help to avoid overfitting---the exact formulation of the
principle is still subject to debate
\citep{OccamsRazor-KDD}, and
several cases have been observed where more complex theories perform
better \citep{DecisionForest,OccamsRazor-Webb,metal:GodShaves}.



Much of this debate focuses on the aspect of predictive accuracy.
When it comes to understandability, the idea that simpler rules are
more comprehensible is typically unchallenged.
A nice counter example is due to \citet{KolmogorovDirections}, who
observed that route directions like \emph{``take every left that doesn't put you on a
prime-numbered highway or street named for a president''} could be 
most compressive but considerably less comprehensive.
%
%
%
Although
\citet{OccamsRazor-KDD} argues in his critical review that it is theoretically and empirically
false to favor the simpler of two models with the same training-set
error on the grounds that this would lead to lower generalization
error, 
he concludes that Occam's Razor is nevertheless relevant for machine learning but should be interpreted
as a preference for more \emph{comprehensible} (rather than
\emph{simple}) models.  Here, the term ''comprehensible'' clearly
does not refer to syntactical length. 

A particular implementation of Occam's razor in machine learning is
the minimum description length \citep[MDL;][]{MDL} or minimum message length \citep[MML\footnote{
	The differences between the two views are irrelevant for our argumentation.};][]{MML} principle
which is an information-theoretic formulation of the principle that
smaller models should be preferred \cite{MDL-Book}. The description length that should
be minimized is the sum of the complexity of the model plus the
complexity of the data encoded given the model. In this way, both the
complexity and the accuracy of a model can be traded off: the
description length of an empty model consists only of the data part,
and it can be compared to the description length of a perfect model,
which does not need additional information to encode the data.
The theoretical foundation of this principle is based on the
Kolmogorov complexity \cite{KolmogorovComplexity}, the essentially
uncomputable length of the smallest model of the data. In practice,
different coding schemes have been developed for encoding models and
data and have, e.g., been used as selection or pruning criteria in decision tree induction \cite{MML-Ockham,MDL-Pruning-2}, inductive rule learning 
\cite{Foil,Ripper,New-MDL} or for pattern evaluation
\cite{Krimp}. 
The ability to compress information has also been proposed as a basis 
for human comprehension and thus forms the backbone of many standard intelligence tests, which aim at detecting patterns in data.
Psychometric artificial intelligence \cite{PsychometricAI} extends this definition
to AI in general. For an extensive treatment of the role of compression in measuring human and machine intelligence we refer the reader to \citet{MeasureOfAllMinds}.
%
%
%

Many works make the assumption that the interpretability of a rule-based model can be measured by measures that relate to the complexity of the model, such as the number of rules or the number conditions.
A maybe prototypical example is the Interpretable Classification Rule Mining (ICRM) algorithm, which ''\emph{is designed to maximize the comprehensibility of the classifier by minimizing the number of rules and the number of conditions}'' via an evolutionary process \citep{InterpretableRuleMining}.
Similarly, \citet{AntMiner-Evaluation} investigate a rule
learner that is able to optimize multiple criteria, and evaluate
it by investigating the Pareto front between accuracy and 
comprehensibility, where the latter is 
measured with
the number of rules.
\citet{InterpretableDecisionSets} 
propose a method for learning rule sets that simultaneously optimizes accuracy and interpretability, where the latter is again measured by several conventional data-driven criteria such as rule overlap, coverage of the rule set, and the number of conditions and rules in the set.  
Most of these works clearly focus on syntactic interpretability.

\subsection{The Bias for Complexity}
\label{sec:bias-complexity}

Even though most systems have a bias toward simpler theories for the
sake of overfitting avoidance and increased accuracy, some rule
learning algorithms strive for more complex rules, and have good
reasons for doing so.
Already
\citet{InductiveLearning} has noted that there are two different kinds
of rules, discriminative and characteristic. 
\emph{Discriminative rules} can quickly discriminate an
object of one category from objects of other categories. A simple
example is the rule 
$$\texttt{elephant :- trunk.}$$
which states that an animal with a trunk is an elephant. This implication
provides a simple but effective rule for recognizing elephants
among all animals. However, it does not provide a very clear picture
on properties of the elements of the target class. For example, from
the above rule, we do not understand that elephants are also very
large and heavy animals with a thick gray skin, tusks and big ears.

\emph{Characteristic rules}, on the other hand, try to capture \emph{all}
properties that are common to the objects of the target class. A rule
for characterizing elephants could be 
$$\texttt{heavy, large, gray, bigEars, tusks, trunk :- elephant.}$$
Note that here the implication sign is reversed: we list all
properties that are implied by the target class, i.e., by an animal
being an elephant. Even though discriminative rules
are easier to comprehend in the syntactic sense, we argue that
characteristic rules are often more
interpretable than discriminative rules from a pragmatic point of view.
For example, in a
customer profiling application, we might prefer to not only list a few
characteristics that discriminate one customer group from the other,
but are interested in all characteristics of each customer group.

The distinction between characteristic and discriminative rule is also reminiscent 
of the distinction between defining and characteristic features of categories.
\citet{SemanticMemory-StructureProcess}\footnote{Many thanks to the anonymous reviewer who pointed us towards this and some of the following works.} argue that both of them are used for
similarity-based assessments of categories to objects, but that only the defining
features are eventually used when similarity-based categorization over all features does not give
a conclusive positive or negative answer.

Characteristic rules are very much related to \emph{formal concept
  analysis} \citep{FCA,FCA-Foundations}. Informally, a concept is
defined by its intent (the description of the concept, i.e., the
conditions of its defining rule) and its extent (the instances that
are covered by these conditions). A \emph{formal concept} is then a\
concept where the extension and the intension are Pareto-maximal, i.e.,
a concept where no conditions can be added without reducing the number
of covered examples. In Michalski's terminology, a formal concept is
both discriminative and characteristic, i.e., a rule where the head is
equivalent to the body.

It is well-known that formal concepts correspond to \emph{closed
  itemsets} in association rule mining, i.e., to maximally specific
itemsets \citep{IcebergLattices}. Closed itemsets have been mined
primarily because they are a unique and compact representative of
equivalence classes of itemsets, which all cover the same instances
\citep{CHARM}. However, while all itemsets in such an equivalence class
are equivalent with respect to their support, they may not be
equivalent with respect to their understandability or
interestingness.

\citet{SD-CHD} introduce \emph{supporting factors} as a means for
complementing the explanation delivered by conventional learned rules.
Essentially, they are additional attributes that are not
part of the learned rule, but nevertheless have very different
distributions with respect to the classes of the application domain. In
a way, enriching a rule with such supporting factors is quite similar
to computing the closure of a rule. In line with the results of
\citet{NB-Rules-Application}, medical experts found that these
supporting factors increase the plausibility of the found rules.

\begin{figure}[t]
\begin{center}
\begin{minipage}[h]{\textwidth}
  \begin{scriptsize}
  					\begin{verbatim}
  					[2160|0]  p :- odor = foul.  
  					[1152|0]  p :- gill-color = buff.  
  					[ 256|0]  p :- odor = pungent.    
  					[ 192|0]  p :- odor = cresote.  
  					[  72|0]  p :- spore-print-color = green.  
  					[  36|0]  p :- stalk-color-below-ring = cinnamon. 
  					[  24|0]  p :- stalk-color-below-ring = scaly.  
  					[   4|0]  p :- cap-surface = grooves.  
  					[   1|0]  p :- cap-shape = conical.  
  					[  16|0]  p :- stalk-color-below-ring = brown, stalk-surface-above-ring = silky.  
  					[   3|0]  p :- habitat = leaves, stalk-color-below-ring = white.
  					  					\end{verbatim}
  				\end{scriptsize}
  			\end{minipage}
  			
  			\medskip
  			(a) using the Laplace heuristic $\hreg_{\textrm{Lap}}$ for refinement
  			
  			\bigskip
  			\begin{minipage}[h]{\textwidth}
  				\begin{scriptsize}
  					\begin{verbatim}
   [2192|0]  p :- veil-color = white, gill-spacing = close, bruises? = no, 
                  ring-number = one, stalk-surface-above-ring = silky.  
   [ 864|0]  p :- veil-color = white, gill-spacing = close, gill-size = narrow, 
                  population = several, stalk-shape = tapering.  
   [ 336|0]  p :- stalk-color-below-ring = white, ring-type = pendant,  
      	           stalk-color-above-ring = white, ring-number = one, 
          	       cap-surface = smooth, stalk-root = bulbous, gill-spacing = close.
   [ 264|0]  p :- stalk-surface-below-ring = smooth, stalk-surface-above-ring = smooth, 
      	           ring-type = pendant, stalk-shape = enlarging, veil-color = white, 
                  gill-size = narrow, bruises? = no. 
   [ 144|0]  p :- stalk-shape = enlarging, stalk-root = bulbous,
                  stalk-color-below-ring = white, ring-number = one.  
   [  72|0]  p :- stalk-shape = enlarging, gill-spacing = close, veil-color = white,
      	           gill-size = broad, spore-print-color = green.  
   [  44|0]  p :- stalk-surface-below-ring = scaly, stalk-root = club.
\end{verbatim}
\end{scriptsize}

\end{minipage}  

\medskip
(b) using the inverted Laplace heuristic $\hinv_{\textrm{Lap}}$ for refinement
\end{center}

\caption{Two decision lists learned for the class \texttt{poisonous} in the
  \data{Mushroom} dataset.}\label{fig:mushroom}
\end{figure}

\citet{jf:ECML-PKDD-14-InvertedHeuristics} introduced so-called
\emph{inverted heuristics} for inductive rule learning. The key idea
behind them is a rather technical observation based on a visualization
of the behavior of rule learning heuristics in coverage space
\cite{jf:MLJ-ROC}, namely that the evaluation of rule refinements is
based on a bottom-up point of view, whereas the refinement process
proceeds top-down, in a general-to-specific fashion. As a remedy, it
was proposed to ``invert'' the point of view, resulting in
heuristics that pay more attention to maintaining high coverage on the
positive examples, whereas  conventional heuristics focus more
on quickly excluding negative examples. 
Somewhat unexpectedly, it turned out that 
this results in longer rules, which resemble characteristic rules instead of the
conventionally learned discriminative rules. For example,
Figure~\ref{fig:mushroom} shows the two decision lists that have been
found for the \data{Mushroom} dataset with the conventional Laplace heuristic 
$\hreg_{\textrm{Lap}}$ (top) and its inverted counterpart $\hinv_{\textrm{Lap}}$
(bottom). Although fewer rules are learned with
$\hinv_{\textrm{Lap}}$, and thus the individual rules are more general
on average, they are also considerably longer. Intuitively, these
rules also look more convincing,
because the first set of rules often only uses a single criterion
(e.g., odor) to discriminate between edible and poisonous
mushrooms. Thus, even though the shorter rules may be more comprehensible in the syntactic sense, the longer rules appear to be more plausible.
\citet{jf:DS-16-ShortRules} and \citet{jf:ESWA} investigated the suitability of
such rules for subgroup discovery, with somewhat
inconclusive results.

\subsection{Conflicting Evidence}
\label{ss:comprmodelsize}

The above-mentioned examples should help to motivate that the complexity of models
may have an effect on the interpretability of a
model. Even in cases where a simpler and a more complex rule covers the
same number of examples, shorter rules are not necessarily more interpretable, at least not when other aspects of interpretability beyond syntactic comprehensibility are considered.
There are a few isolated empirical studies that add to this picture. However, 
the results on the relation between the size of representation and interpretability are limited and conflicting, partly because different aspects of 
interpretability are not clearly discriminated.

\paragraph{Larger Models are Less Interpretable.}
\citet{huysmans2011empirical} were among the first that actually tried to empirically validate the often implicitly made claim that smaller models are more interpretable. 
In particular, they related increased complexity to measurable events such as 
a decrease in answer accuracy, an increase in answer time, and a decrease in confidence.
From this, they concluded that smaller models tend to be more interpretable, proposing that there is a certain complexity threshold that limits the practical utility of a model. However, they also noted that in parts of their study, the correlation of model complexity with utility was less pronounced.
The study also does not report whether
the participants of their study had any domain knowledge relating to the used
data, so  
that it cannot be ruled out that the obtained result
was caused by  lack of domain knowledge.\footnote{The lack of domain knowledge was hypothesized to account for differences observed in another study by \cite{User-ModelUnderstandability},
which we discuss in more detail below.}
 A similar study was later conducted by \citet{DecisionTreeComprehensibility}, who found a clear relationship between model complexity and interpretability in decision trees.
 
In most previous works, interpretability was interpreted
 in the sense of  
 syntactic comprehensibility, i.e., the pragmatic or epistemic aspects of interpretability were not addressed.

\paragraph{Larger Models are More Interpretable.}
A direct evaluation of the perceived interpretability of classification models has been performed by \citet{User-ModelUnderstandability}. 
They elicited preferences on pairs of models which were generated from two UCI datasets: \data{Labor} and \data{Contact Lenses}. What is unique to this study is that the analysis took into account the participants' estimated knowledge about the domain of each of the datasets.  On \data{Labor}, they were expected to have good domain knowledge but not so for \data{Contact Lenses}. The study was performed with 100 students and involved several decision tree induction algorithms (\alg{J48}, \alg{RIDOR}, \alg{ID3}) as well as rule learners (\alg{PRISM}, \alg{Rep}, \alg{JRip}). It was found that \emph{larger models} were considered as \emph{more comprehensible} than smaller models on the \data{Labor} dataset whereas the users showed the opposite preference for \data{Contact Lenses}.  
\citet{User-ModelUnderstandability}
explain the  discrepancy  with the lack of prior knowledge for \data{Contact Lenses}, which makes it harder to understand complex models, whereas in the case of \data{Labor}, 
\emph{``\ldots the  larger  or  more  complex classifiers did  not  diminish  the  understanding of  the decision process,  but  may  have  even increased  it  through  providing  more  steps  and  including  more attributes for each  decision  step.''}
%
%
%
In an earlier study,
\citet{NB-Rules-Application} found that medical experts rejected rules
learned by a decision tree algorithm because they found them to be too
short. Instead, they preferred explanations that were derived from a
Na\"ive Bayes classifier, which essentially showed weights for all
attributes, structured into confirming and rejecting attributes.

\medskip
To some extent, the results may appear to be inconclusive because the different
studies do not clearly discriminate between different aspects of interpretability. 
Most of the results that report that simpler models are more interpretable
refer to syntactic interpretability, whereas, e.g.,  \citet{User-ModelUnderstandability} tackle epistemic interpretability
by taking the users' prior knowledge into account. Similarly, the study of 
\citet{NB-Rules-Application} has aspects
of epistemic interpretability, in that ''too short'' explanations 
contradict the experts' experience.
pragmatic interpretability of the models has not been explicitly addressed, nor are we aware of any studies that explicitly relate plausibility to model complexity.

\subsection{Experiment 1: Are Shorter Rules More Plausible?}
\label{sec:shorter}

Motivated by the somewhat inconclusive evidence in previous works on interpretability and complexity, we set up a crowdsourcing experiment that specifically focuses on the aspect of plausibility. In this and the experiments reported in subsequent sections, the basic experimental setup follows the one discussed in Section~\ref{sec:crowd}. Here, we only note task-specific aspects.

\begin{figure}[tb]
	\centering
	\resizebox{\textwidth}{!}{
		\fbox{\includegraphics{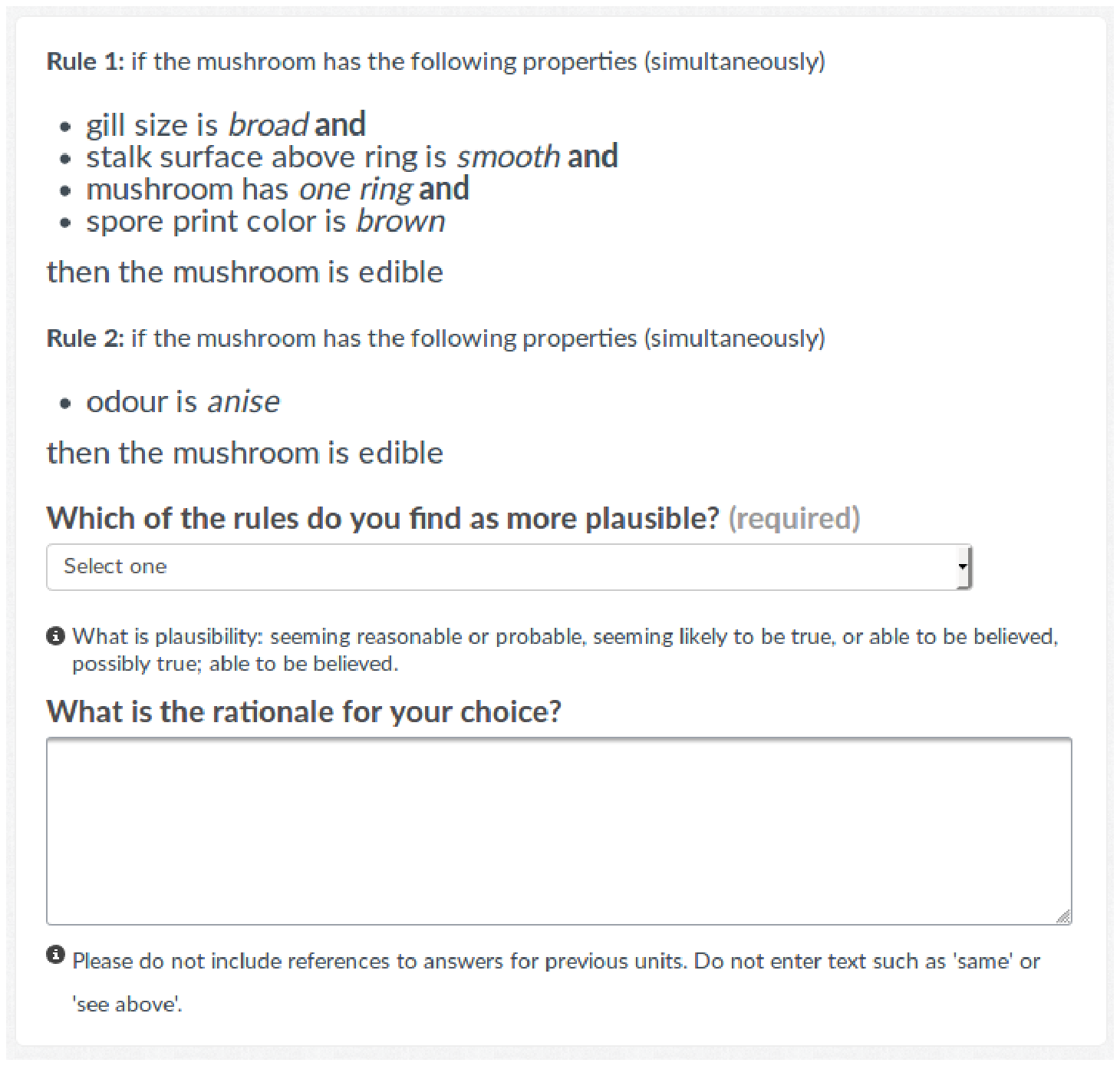}}}
	\caption{Example rule pair used in experiments 1--3. For Experiment 3, the description of the rule also contained values of confidence and support, formatted as shown in Figure~\ref{fig:supp-conf}.}
	\label{fig:RulePairScreenshot}
\end{figure}

\paragraph{Material.}
The questionnaires presented pairs of rules as described in section~\ref{sec:rule-generation}, and asked the participants to give a) judgment which
rule in each pair is more preferred and b) optionally a textual
explanation for the judgment.
%
 A sample question is shown in
Figure~\ref{fig:RulePairScreenshot}.
The judgments were elicited using a drop down box, where the participants could choose from the following five options: \emph{``Rule~1 (strong preference)''}, \emph{``Rule~1 (weak preference)''}, \emph{``No preference''}, \emph{``Rule~2 (weak preference)''}, \emph{``Rule~2 (strong preference)''}. As shown in Figure~\ref{fig:RulePairScreenshot}, the definition  of plausibility was accessible to participants at all times, since it was featured below the drop-down box.
As optional input, the workers could provide a textual explanation of their reasoning behind the assigned preference, which we informally evaluated but which is not further considered in the analyses reported in this paper.


\paragraph{Participants.}
The number of judgments per rule pair 
for this experiment was 5 for the \data{Traffic}, \data{Quality}, and \data{Movies} datasets. The \data{Mushroom} dataset had only 10 rule pairs, therefore we opted to collect 25 judgments for each rule pair in this dataset.

\paragraph{Results.}

Table~\ref{tbl:cf-rulelength-statistics} summarizes the results of this crowdsourcing experiment. In total, we collected $1002$ responses, which is on average 6.3 judgments for each of the $158$ rule pairs.
%
On two of the datasets, \data{Quality} and \data{Mushroom}, there was a strong, statistically significant \emph{positive} correlation between rule length and the observed plausibility of the rule, i.e., longer rules were preferred.
In the other two datasets, \data{Traffic} and \data{Movies}, no significant difference could be observed in either way.

In any case, these results show that there is no \emph{negative}
correlation between rule length and plausibility. In fact, in two of
the four datasets, we even observed a positive correlation, meaning
that in these cases longer rules were preferred.

\begin{table}[h!]
	\caption{Rule-length experiment statistics. \emph{pairs} refers to the distinct number of rule pairs,  \emph{judg} to the number of trusted judgments, the quiz failure rate \emph{qfr} to the percentage of participants that did not pass the initial quiz as reported by the CrowdFlower dashboard, \emph{part} to the number of trusted distinct workers, $\tau$ and $\rho$ to the observed correlation values with $p$-values in parentheses.}
	\medskip
	\centering
	\begin{tabular}{lccccrcrc}
		\toprule
		& \multicolumn{1}{l}{pairs} & \multicolumn{1}{c}{judg} & \multicolumn{1}{c}{qfr} & \multicolumn{1}{c}{part} & \multicolumn{2}{c}{Kendall's $\tau$} & \multicolumn{2}{c}{Spearman's $\rho$} \\ \midrule
		\data{Traffic} & 80 & 408 & 11 & 93 & 0.05 &\scriptsize{(0.226)} & 0.06 &\scriptsize{(0.230)} \\ 
		\data{Quality} & 36 & 184 & 11 & 41 & $\mathbf{0.20}$ &\scriptsize{(0.002)} & $\mathbf{0.23}$ &\scriptsize{(0.002)} \\ 
		\data{Movies} & 32 & 160 & 5 & 40 & -0.01& \scriptsize{(0.837)} & -0.02 &\scriptsize{(0.828)} \\ 
		\data{Mushrooms} & 10 & 250 & 13 & 84 & $\mathbf{0.37}$ & \scriptsize{(0.000)} & $\mathbf{0.45}$ &\scriptsize{(0.000)} \\ \midrule
		total & 158 & 1002 & 11 & 258 &  &  & &\\ \bottomrule
	\end{tabular}
	\label{tbl:cf-rulelength-statistics}
\end{table}

\section{The Conjunction Fallacy}
\label{sec:conjunction-fallacy}

Human-perceived plausibility of a hypothesis has been extensively studied in cognitive science.  The best-known cognitive phenomenon related to our focus area of the influence of the number of conditions in a rule on its plausibility is the \emph{conjunctive fallacy}. This fallacy falls into the research program on cognitive biases and heuristics carried out by Amos Tversky and Daniel Kahneman since the 1970s. 
The outcome of this research program can be succinctly summarized by a quotation from Kahneman's Nobel Prize lecture 
at Stockholm University on December 8, 2002:
\begin{quote}
\it  ``$\ldots$, it is safe to assume that similarity is more accessible than probability,  that  changes  are  more  accessible  than  absolute values,  that  averages  are  more  accessible  than  sums,  and that the accessibility of a rule of logic or statistics can be temporarily increased by a reminder.'' \rm \hfill \citep{kaheman2003perspective}
\end{quote}

In this section, we will briefly review some aspects of this program, highlighting those that seem to be important for inductive rule learning. For a more thorough review we refer to \citet{JudgementUncertainty} and \citet{HeuristicsAndBiases}, a more recent, very accessible introduction can be found in \citet{ThinkingFastSlow}.

\begin{figure}[t]

\begin{Verbatim}[frame=single]

	Linda is 31 years old, single, outspoken, and very bright.
	She majored in philosophy. As a student, she was deeply 
	concerned with issues of discrimination and social justice, 
	and also participated in anti-nuclear demonstrations.
	
	Which is more probable?
	
	(a) Linda is a bank teller.
	(b) Linda is a bank teller and is active in the
	    feminist movement.
	    
\end{Verbatim}      

\caption{The Linda problem \citep{tversky1983extensional}.}
\label{fig:Linda}
\end{figure}

\subsection{The Linda Problem}
\label{ss:concfallacy}
\label{sec:linda}

The conjunctive fallacy is in the literature often defined via the ``Linda'' problem. In this problem, participants are asked whether they consider it more plausible that a person Linda is more likely to be (a) a bank teller or (b) a feminist bank teller (Figure~\ref{fig:Linda}).
\citet{tversky1983extensional}  report that based on the provided characteristics of Linda, 85\% of the participants indicate (b) as the more probable option. This was essentially confirmed in by various independent studies, even though the actual proportions may vary.
In particular, similar results could be observed
across multiple settings (hypothetical scenarios, real-life domains), as well as for various kinds of participants (university students, children, experts, as well as statistically sophisticated individuals) \citep{tentori2012conjunction}. 

However, it is easy to see that
the preference for (b)
is in  conflict with elementary laws of probabilities.
Essentially, in this example, participants are asked
to compare conditional probabilities  $\Pr(F \wedge B \mid L)$ and $\Pr(B\mid L)$, where $B$ refers to ``bank teller'', $F$ to  ``active in feminist movement'' and $L$ to the description of Linda. 
Of course,
the probability of a conjunction, $\Pr(A \wedge B)$, cannot exceed the probability of its constituents, $\Pr(A)$ and $\Pr(B)$ \citep{tversky1983extensional}. In other words, as it always holds for the Linda problem that  $\Pr(F \wedge B \mid L) \leq \Pr(B \mid L)$, the preference for alternative $F\wedge B$ (option (b) in Figure~\ref{fig:Linda}) is a logical fallacy.

\subsection{The Representativeness Heuristic}
\label{sec:repres}
According to \citet{tversky1983extensional},
the results of the conjunctive fallacy experiments manifest
that \emph{``a conjunction can be more representative than one of its constituents''}.
It is a symptom of a more general phenomenon, namely that people
tend to overestimate the probabilities of representative events and underestimate those of less representative ones. The reason  is attributed to the application of the \emph{representativeness heuristic}. This heuristic provides humans with means for assessing a probability of an uncertain event.
According to the representativeness heuristic, the probability that an object A belongs to a class B is evaluated \emph{``by the degree to which A is representative of B, that is by the degree to which A resembles B''} \citep{Tversky27091974}.

This heuristic relates to the tendency to make judgments based on similarity, based on a rule ``like goes with like''. 
According to \citet{gilovich2002like}, the representativeness heuristic can be held accountable for a number of widely held false and pseudo-scientific beliefs, including those in astrology or graphology.\footnote{\citet{gilovich2002like} gives the following example: resemblance of the physical appearance of the sign, such as crab, is related in astrology with personal traits, such as appearing tough on the outside. For graphology, the following example is given: handwriting to the left is used to indicate that the person is holding something back.} It can also inhibit valid beliefs that do not meet the requirements of resemblance.


A related phenomenon is that people often tend to misinterpret the meaning 
of the logical connective ``and''. 
\citet{Hertwig2008conjunction} hypothesized that the conjunctive fallacy could be caused by ``a  misunderstanding about conjunction'', i.e., by a different interpretation of ``probability'' and ``and''  by the participants than assumed by the experimenters.
They discussed that ``and'' in natural language can express several relationships, including temporal order, causal relationship,  and most importantly, can also indicate a collection of sets instead of their intersection. For example, the sentence ``He invited friends and colleagues to the party'' does not mean that all people at the party were both colleagues and friends.
%
%
According to \citet{sides2002reality}, ``and'' ceases to be ambiguous when it is used to connect propositions  rather than categories. The authors give the following example of a sentence which is not prone to misunderstanding: ``IBM stock will rise tomorrow and Disney stock will fall tomorrow''.
Similar wording of rule learning results may be, despite its verbosity, preferred.
We further conjecture that representations that visually express the semantics of ``and'' such as decision trees may be preferred over rules, which do not provide such visual guidance. 

\subsection{Experiment 2: Misunderstanding of ``and'' in Inductively Learned Rules}
Given its omnipresence in rule learning results, it is vital to assess to what degree the ``and'' connective is misunderstood when rule learning results are interpreted. 
%
%
In order to gauge the effect of the conjunctive fallacy, we carried out a separate set of crowdsourcing tasks,  
To control for misunderstanding of ``and'', the group of workers approached in Experiment 2 additionally received \emph{intersection} test questions which 
were intended to ensure that all participants understand  the \emph{and} conjunction the same way it is defined in the probability calculus. 
In order to correctly answer these, the respondent had to realize that the antecedent of one of the rules contains mutually exclusive conditions. The correct answer was a weak or strong preference for rule which did not contain the mutually exclusive conditions. 



\paragraph{Material.}
The participants were presented with the same rule pairs as in Experiment~1 (Group~1).
The difference between Experiment~1 and Experiment~2 was only one manipulation: instructions in Experiment~2  additionally contained the intersection test questions, not present in Experiment~1. We refer to the participants that received these test questions as Group~2.

\paragraph{Participants.} Same as for Experiment~1 described earlier.
There was one small change for the \data{Mushroom} dataset, where for economical constraints we collected 15 judgments for each rule pair within Experiment~2, instead of 25 collected in Experiment~1.

\begin{table}[t]
	\caption{Effect of intersection test questions that are meant to ensure that participants understand the logical semantics of ''and''.
		\emph{pairs} refers to the distinct number of rule pairs,  \emph{judg} to the number of trusted judgments, the quiz failure rate \emph{qfr} to the percentage of workers that did not pass the initial quiz as reported by the CrowdFlower dashboard, \emph{part} to the number of trusted distinct workers, and $\tau$ to the observed correlation values with $p$-values in parentheses.}
	\label{tab:intersection}
	\medskip
	\centering
\begin{tabular}{ll|lrrrl|lrrrl}
\toprule
 \multicolumn{2}{c}{} & \multicolumn{5}{c}{Group 1: w/o int. test questions}  &  \multicolumn{5}{c}{Group 2: with int. test questions}  \\ 
 
dataset & \multicolumn{1}{l}{pairs} & judg & qfr & part & \multicolumn{2}{c}{Kendall's $\tau$} &  judg & qfr & part & \multicolumn{2}{c}{Kendall's $\tau$}\\ 
\midrule
\data{Quality} & 36 & 184 & 11 & 41 & $\mathbf{0.20}$ &  \scriptsize{(0.002)} & 180 & 31 & 45 & -0.03 & \scriptsize{(0.624)} \\ 
\data{Mushroom} & 10 & 250 & 13 & 84 & $\mathbf{0.37}$ &  \scriptsize{(0.000)} & 150 & 44 & 54 & $\mathbf{0.28}$ & \scriptsize{(0.000)}  \\ 
\bottomrule
\end{tabular}
\label{}
\end{table}

\paragraph{Results.}
We state the following proposition: \emph{The effect of higher perceived interpretability of longer rules disappears when it is ensured that participants understand the semantics of the ``and'' conjunction}. The corresponding null hypothesis is that the correlation between rule length and plausibility
is no longer statistically significantly different from zero for participants that successfully completed the intersection test questions (Group 2).
We focus on the analysis on \data{Mushroom} and \data{Quality} datasets on which we had initially observed a higher plausibility of longer rules.

The results 
presented in Table~\ref{tab:intersection}
show that the correlation coefficient is still statistically significantly different from zero for the \data{Mushroom} dataset with Kendall's $\tau$ at 0.28 ($p<0.0001$), but not for the \data{Quality} dataset,  which has $\tau$  not different from zero at $p<0.05$ (albeit at a much higher variance).
This suggests that at least on the \data{Mushroom} dataset, there are other factors apart from ``misunderstanding of and'' that cause longer rules to be perceived as more plausible. We will take a look at some possible causes in the following sections.


\section{Insensitivity to Sample Size}
\label{sec:insensitivity}

In the previous sections, we have motivated that rule length is by itself not an indicator for the plausibility of a rule if other factors such as the support and the confidence of the rule are equal. In this and following sections, we will discuss the influence of these and a few alternative factors, partly motivated by results from the psychological literature. The goal is to motivate some directions for future research on the interpretability and plausibility of learned concepts.

\subsection{Support and Confidence}

In the terminology used within the scope of cognitive science \citep{griffin1992weighing}, confidence corresponds to the \emph{strength} of the evidence and support to the \emph{weight} of the evidence.
Results in cognitive science for the strength and weight of evidence suggest that the weight is systematically undervalued while the strength is overvalued.  
According to \citet{Camerer1992}, this was, e.g., already mentioned by \citet{keynes1922treatise} who drew attention to the problem of balancing the likelihood of the judgment and the weight of the evidence in the assessed likelihood. 
In particular, \citet{BeliefInSmallNumbers} have argued that human analysts are unable to appreciate the reduction of variance and the corresponding increase in reliability of the confidence estimate with increasing values of support.
This bias is known as \emph{insensitivity to sample size}, and essentially describes the human tendency to  neglect the following two principles: a) more variance is likely to occur in smaller samples, b) larger samples provide less variance and better evidence.
Thus,
people underestimate the increased benefit of higher robustness of estimates made on a larger sample.

In the previous experiments, we controlled the rules selected into the pairs so they mostly had identical or nearly identical confidence and support. Furthermore, the confidence and support values of the shown rules were \emph{not} revealed to the participants during the experiments. 
However, in real situations, rules in the output of inductive rule learning have varying quality, which is communicated mainly by the values of confidence and support.  
Given that longer rules can fit the data better, they tend to be higher on confidence and lower on support.
This implies that if confronted with two rules of different length, where the longer has a higher confidence and the shorter a higher support, the analyst may prefer the longer rule with higher confidence (all other factors equal).  
%
These deliberations lead us to the following proposition:
\emph{When both confidence and support are explicitly revealed, confidence but not support will positively increase rule plausibility.}

\subsection{Experiment 3: Is rule confidence perceived as more important than support?}

We aim to evaluate the effect of explicitly revealed confidence (strength) and support (weight) on rule preference.  
In order to gauge the effect of rule quality measures confidence and support, we
performed an additional experiment.

\begin{figure}[t]
	\centering\fbox{%
		\begin{minipage}{0.95\textwidth}
			\emph{Rule 1}: if the movie falls into all of the following group(s) (simultaneously) \emph{English-language Films} then the movie is rated as \textbf{bad}
			
			\smallskip
			\textbf{Additional Information:} 
			if the movie falls into all of the following group(s) (simultaneously) Englishlanguage Films then the movie is rated as \textbf{bad}
			
			In our data, there are \textbf{995} movies which match the conditions of this rule. Out of these \textbf{518} are predicted correctly as having bad rating. The confidence of the rule is \textbf{52\%}. 
			
			In other words, out of the \textbf{995} movies that match all the conditions of the rule, the number of movies that are rated as \textbf{bad} as predicted by the rule is 518. The rule thus predicts correctly the rating in  518/995=52 percent of cases.
			
			\medskip
			\emph{Rule 2}: if the movie falls into all of the following group(s) (simultaneously) 
			\emph{Films Released In 2010} and \emph{English-language Films} then the movie is rated as \textbf{bad}
			
			\smallskip
			\textbf{Additional Information:} 
			In our data, there are \textbf{55} movies which match the conditions of this rule. Out of these \textbf{29} are predicted correctly as having bad rating. The confidence of the rule is \textbf{53\%}. 
			
			In other words, out of the \textbf{55} movies that match all the conditions of the rule, the number of movies that are rated as \textbf{bad} as predicted by the rule is 29. The rule thus predicts correctly the rating in  29/55=53 percent of cases.
		\end{minipage}
	}
	\caption{Rule pair including the additional information on support and confidence.}
	\label{fig:supp-conf}
\end{figure}

\paragraph{Material.}
The participants were presented with rule pairs like in the previous two experiments. We used only rule pairs from 
the \data{Movies} dataset, where the differences in confidence and support between the rules in the pairs were largest. 
The only difference in the setup between Experiment~1 and Experiment~3  was that participants now
also received information about the number of correctly and incorrectly covered instances for each rule, along with the support and confidence values.  
Figure~\ref{fig:supp-conf} shows an
example of this additional information provided to the participants.
Workers that received this extra information are referred to as Group~3.

\paragraph{Participants.}
This setup was the same as for the preceding two experiments.


\begin{table}[t]
	\caption{Kendall's $\tau$ on the \data{Movies} dataset with (Group 1) and without (Group 2) additional information about the number of covered good and bad examples.
\emph{pairs} refers to the distinct number of rule pairs,  \emph{judg} to the number of trusted judgments, the quiz failure rate \emph{qfr} to the percentage of workers that did not pass the initial quiz as reported by the CrowdFlower dashboard, \emph{part} to the number of trusted distinct workers, and $\rho$ to the observed correlation values with $p$-values in parentheses.}		
	
	\label{tab:supp-conf}
	\medskip
	\centering
	\begin{tabular}{lr|rrrrl|rrrrl}
		\toprule
		\multicolumn{1}{c}{}& \multicolumn{1}{c}{} & \multicolumn{5}{c}{Group 1} & \multicolumn{5}{c}{Group 3} \\
		\multicolumn{1}{c}{} & \multicolumn{1}{c}{} & \multicolumn{5}{c}{Without information} & \multicolumn{5}{c}{With information} \\
		\multicolumn{1}{l}{measure} & \multicolumn{1}{c}{pairs} & judg & qfr & part & \multicolumn{2}{c}{Kendall's $\tau$} &  judg & qfr & part & \multicolumn{2}{c}{Kendall's $\tau$}\\
		\midrule
		Support    & \multirow{ 2}{*}{32} & \multirow{ 2}{*}{160} & \multirow{ 2}{*}{5} & \multirow{ 2}{*}{40} & $-0.07$ & \scriptsize{($0.402$)} &  \multirow{ 2}{*}{160
		} & \multirow{ 2}{*}{5}  & \multirow{ 2}{*}{40} & $-0.08$ &\scriptsize{($0.361$)} \\
		Confidence &  & & & & $0.00$ & \scriptsize{($0.938$)} &  & & & \multicolumn{1}{r}{$\mathbf{ 0.24}$} & \scriptsize{($0.000$)} \\
		\bottomrule
	\end{tabular}
\end{table}

\paragraph{Results.}
Table~\ref{tab:supp-conf} shows the correlations of the rule quality measures confidence and support with plausibility. 
It can be seen that there is a relation
to confidence but not to support, even though both were explicitly present in descriptions of rules for Group 3.
Thus, our result supports the hypothesis that insensitivity to sample size effect is applicable to the interpretation of inductively learned rules. In other words, when both confidence and support are stated, confidence positively affects the preference for a rule whereas support tends to have no impact.

The results also show that the relationship between revealed rule confidence and plausibility is causal. This follows from confidence not being correlated with plausibility in the original experiment (Group 1 in Figure~\ref{tab:supp-conf}), which differed only via the absence of the  explicitly revealed information about rule quality. While such conclusion is intuitive, to our knowledge it has not yet been empirically confirmed before.
%
%

\section{Relevance of Conditions in Rule}
\label{sec:relevance}

An obvious factor that can determine the perceived plausibility of a proposed rule is how relevant it appears to be.
Of course, rules that contain more relevant conditions will be considered to be more acceptable. One way of measuring this could be in the strength of the connection between the condition (or a conjunction of conditions) with the conclusion. However, in our crowdsourcing experiments we only showed sets of conditions that are equally relevant in the sense that their conjunction covers about the same number of examples in the shown rules or that the rules have a similar strength of connection. Nevertheless, the perceived or subjective relevance of a condition may be different for different users. 

There are several cognitive biases that can distort the correlation between the relevance of conditions and the the judgment of plausibility. One of the most recently discovered is the \emph{weak evidence effect}, according to which evidence in favor of an outcome can actually decrease the probability that a person assigns to it. In an experiment in the area of forensic science reported by \citet{martire2013expression}, it was shown that participants presented with evidence weakly supporting guilt tended to ``invert'' the evidence, thereby counterintuitively reducing their belief in the guilt of the accused.

\enlargethispage*{12pt}
\subsection{Attribute and Literal Relevance}

In order to analyze the effect of relevance in the rule learning domain,  we decided 
to enrich our input data with two supporting crowdsourcing tasks, which aimed at collecting judgments of attribute and literal relevance.

\paragraph{Attribute Relevance.}
Attribute relevance corresponds to human perception of the ability of a specific attribute to predict values of the attribute in rule consequent. 
%
%
%
%
%
%
%
%
For example, in the \data{Movies} data, the release date of a film may be perceived as less relevant for determining the quality of a film than its language.  
%
Attribute relevance 
also reflects
a level of recognition of the explanatory attribute (cf.\ also Section~\ref{sec:recognition}), which is a prerequisite to determining the level of association with the target attribute.  As an example of a specific attribute that may not be recognized consider ``Sound Mix'' for a movie rating problem. This would contrast with attributes such as ``Oscar winner'' or ``year of release'', which are equally well recognized, but clearly associated to a different degree with the target.

\paragraph{Literal Relevance.}
Literal relevance goes one step further than attribute relevance by measuring human perception of the ability of a specific condition to predict a specific value of the attribute in the rule consequent.
%
%
%
It should be noted that we consider the literal relevance to also embed 
attribute relevance to some extent.
For example, 
the literal (``film released in 2001'') conveys also the attribute (``year of release''). However, in addition to the attribute name, literal also conveys a specific value, which may not be 
recognized by itself. This again raises the problem of recognition as a prerequisite to association. 


\begin{figure}[t]
	\begin{minipage}[t]{0.49\textwidth}	
		\begin{scriptsize}
			\begin{Verbatim}[frame=single]
We kindly ask you to assist us in an 
experiment that will help researchers 
understand which properties influence  
mushroom being considered as 
poisonous/edible.

Example task follows:

Property: Cap shape

Possible values: bell, conical, convex, 
                 flat, knobbed, sunken
			
What is the relevance of the property 
given above for determining whether a 
mushroom is edible or poisonous?

Give a judgment on a 10 point scale:

     1 = Completely irrelevant
    10 = Very relevant
\end{Verbatim}
			%
			%
			%
			%
		\end{scriptsize}
		\caption{Attribute relevance question for \data{Mushroom}.}
		\label{fig:cf-mushroom-attrel}
	\end{minipage}
	\hfill
	\begin{minipage}[t]{0.49\textwidth}
		\begin{scriptsize}
			\begin{Verbatim}[frame=single]
We kindly ask you to assist us in an 
experiment that will help researchers 
understand  which factors can influence 
movie ratings.


Example task follows:

Condition: Academy Award Winner or Nominee

The condition listed above will 
contribute to a movie being rated as:

    Good (Strong influence)
    Good (Weak influence)
    No influence
    Bad (Weak influence)
    Bad (Strong influence)
	
Select one option.

\end{Verbatim}
			%
			%
			%
			%
			%
		\end{scriptsize}
		\caption{Literal relevance test question for \data{Movies}.}
		\label{fig:LiteralEx}
	\end{minipage}
\end{figure}

\subsection{Experiment 4: Influence of Attribute and Literal Relevance}
The experiments
were performed similarly as the previous ones using crowdsourcing.
Since the relevance experiments did not elicit preferences for rule pairs, there are multiple differences from the setup described earlier.
We summarize the experiments in the following, but refer the reader to \citet{kliegr2017thesis} for additional details.

\paragraph{Material.}
The data collected within Experiments 1--3 were enriched with variables denoting the relevance of attributes and literals of the individual rules. Given that in Experiments 1--3 plausibility was elicited for rule pairs, the variables representing relevance were computed as differences of values obtained for the rules in the pair. 

\pagebreak
Each rule pair was enriched with four\footnote{We initially also experimented with computing several other enrichment variables not reported here (derived from label length and depth in the taxonomy, using ratios instead of differences, and using minimum in addition to average and maximum). For these variables, we either did not obtain statistically significant results, or the interpretation was not intuitive, therefore we do not report these additional evaluations here.} 
variables according to the following pattern: ``[Literal$\vert$Attribute]\linebreak Rel[Avg$\vert$Max]$\Delta$''. To compute the enrichment variable, the value of the relevance metric for the second rule in the pair (r2) was subtracted from the value for the first rule (r1).
For example, 
\begin{equation}
 LiteralRelAvg\Delta= LiteralRelAvg(r1)-LiteralRelAvg(r2),
\end{equation}
 where \emph{LiteralRelAvg(r1)}, \emph{LiteralRelAvg(r2)} represent the average relevance of literals (conditions) present in the antecedent of rule r1 (r2) in the pair.

The attribute relevance experiments were prepared for the \data{Mushroom} and \data{Traffic} datasets.
An example wording of the attribute relevance elicitation task for the Mushroom dataset is shown in Figure~\ref{fig:cf-mushroom-attrel}.
%
%
An example wording of the literal relevance elicitation task for the Movies dataset is shown in Figure~\ref{fig:LiteralEx}.
%
In this case, there was a small difference in setup between the experiments on LOD datasets and the \data{Mushroom} dataset. The latter task did contain links to Wikipedia for individual literals as these were directly available from the underlying dataset. For the \data{Mushroom} dataset no such links were available and thus these were not included in the task.

\begin{table}[t]
	\caption{Attribute and Literal Relevance (Group 1, Kendall's $\tau$). Column \emph{att} refers to number of distinct attributes, \emph{lit} to number of distinct literals (attribute-value pairs), \emph{judg} to the number of trusted judgments,  \emph{excl} to the percentage of workers that were not trusted on the basis of giving justifications shorter than 11 characters, and \emph{part} to the number of trusted distinct workers.}		
	\label{tab:relevance}
	\medskip
	\centering
	\begin{tabular}{lrrrrrlrl}
		\toprule
		\multicolumn{9}{c}{Attribute relevance} \\
		
		Dataset & att&	judg&	excl	& part 	& \multicolumn{2}{c}{Avg} & \multicolumn{2}{c}{Max} \\ \midrule
		\data{Traffic} & 14 &	35	& 70&	6
		&  $0.01$ &  \scriptsize{($0.757$)} &  $0.00$ & \scriptsize{($0.983$)} \\
		\data{Mushroom} & 10	& 92	& 66 &	31
		&  $-0.11$ &  \scriptsize{($0.018$)} &  $\mathbf{0.27}$ & \scriptsize{($0.000$)} \\
		\midrule
		\multicolumn{9}{c}{Literal relevance} \\
		
		Dataset & lit&	judg&	excl	& part & \multicolumn{2}{c}{Avg} & \multicolumn{2}{c}{Max} \\\midrule
		\data{Quality} & 33 &	165&	40&	45
		
		&  $\mathbf{0.29}$ &  \scriptsize{($0.000$)} &  $\mathbf{0.31}$ & \scriptsize{($0.000$)} \\
		\data{Movies} & 30&	150&	19&	40
		& $0.15$ &  \scriptsize{($0.012$)} & $\mathbf{0.22}$ & \scriptsize{($0.000$)} \\
		\data{Traffic} & 58&	290&	40&	75
		& $0.04$ &  \scriptsize{($0.311$)} &  $0.01$ & \scriptsize{($0.797$)}\\
		\data{Mushroom} & 34&170&16&42
		& $\mathbf{-0.19}$ &  \scriptsize{($0.000$)} &  $0.11$ & \scriptsize{($0.037$)} \\
		\bottomrule
	\end{tabular}
\end{table}

\paragraph{Results.}
Table~\ref{tab:relevance} shows the correlations between plausibility and the added variables representing attribute and literal relevance on the data collected for Group 1 from the previous experiments. The results confirm that literal relevance has a strong correlation with the judgment of the plausibility of a rule. A rule which contained (subjectively) more relevant literals than the second rule in the pair was more likely to be evaluated favorably than rules that do not contain such conditions. This pattern was found valid with varying levels of statistical significance across all evaluation setups in Table~\ref{tab:relevance}, except for the average relevance in the smallest \data{Mushroom} dataset.

Note that the effect is strongest for the maximum relevance, which means that it is not necessary that all the literals are deemed important, but it suffices if a few (or even a single) condition is considered to be relevant. 
\data{Traffic} was the only dataset where such effects could not be observed, but this may have to do with the fact that the used attributes (mostly geographic regions) strongly correlate with traffic accidents but do not show a causal relationship. 
The examination of the relation between the objective relevance of conditions in a rule and their impact on the subjective perception of the rule is an interesting yet challenging area of further study. The perception can be influenced by multiple cognitive phenomena, such as the weak evidence effect. 


\section{Recognition Heuristic}
\label{sec:recognition}

The recognition heuristic \citep{goldstein1999recognition,goldstein2002models} is the best-known of the fast and frugal heuristics that have been popularized in several books, such as \citet{Gigerenzer-SimpleHeuristics,Gigerenzer-Heuristics,Gigerenzer-SimplyRational}.
It essentially states that when you compare two objects according to some criterion that you cannot directly evaluate, and
	''\textit{one of two objects is recognized and the other is not, then infer that the recognized object has the higher value with respect to the criterion.}''
Note that this is independent of the criterion that should be maximized, it only depends on whether there is an assumed positive correlation with the recognition value of the object.
For example, if asked whether Hong Kong or Chongqing is the larger city, people tend to pick Hong Kong because it is better known (at least in the western hemisphere), even though Chongqing has about four times as many inhabitants. Thus, it may be viewed as being closely associated to relevance, where, in the absence of knowledge about a fact, the city's relevance is estimated by how well it is recognized.



The recognition heuristic can manifest itself as a preference for rules containing a recognized literal or attribute in the antecedent of the rule. Since the odds that a literal will be recognized increase with the length of the rule, it seems plausible that the recognition heuristic generally increases the preference for longer rules. 
One could argue that for longer rules, the odds of occurrence of an unrecognized literal will also increase. The counterargument is the empirical finding that---under time pressure---analysts assign recognized objects a higher value than to unrecognized objects. This happens also in situations when recognition is a poor cue \citep{pachur2006psychology}. 



\subsection{Experiment 5: Modeling Recognition Heuristic using PageRank}
In an attempt to measure representativeness, we resort to measuring the centrality of a concept using its PageRank \cite{PageRank} in a knowledge graph. 
In three of our datasets, the literals correspond to Wikipedia articles, which allowed us to use PageRank computed from the Wikipedia connection graph for these literals.  Similarly as for the previous experiment, each rule pair was enriched with two additional variables corresponding to the difference in the average and maximum PageRank associated with literals in the rules in the pair. We refer the reader to \citet{kliegr2017thesis} for additional details regarding the experimental setup.
%
%

\begin{table}[t]
  \caption{Correlation of PageRank in the knowledge graph with plausibility (Group 1, Kendall's $\tau$). Column \emph{lit} refers to number of distinct literals (attribute-value pairs), \emph{judg} to the number of trusted judgments,  \emph{qfr} to the percentage of non-trusted workers, and \emph{part} to the number of trusted distinct workers.}
  \label{tab:pagerank}
  \medskip
\centering
  \begin{tabular}{lrrrrrcrc}
    \toprule
    Dataset & lit	& judg &	qfr & part
 & \multicolumn{2}{c}{Avg} & \multicolumn{2}{c}{Max} \\
    \midrule
    
    \data{Quality} & 33	 & 165 &40 &45 &  $0.01$ &  \scriptsize{($0.882$)} &  $0.07$ & \scriptsize{($0.213$)} \\
    \data{Movies} & 30 & 150 &19 & 40 &  \scriptsize{($0.051$)} & $-0.07$ & \scriptsize{($0.275$)} \\
    \data{Traffic} & 58	& 290 &	40 & 75  & $0.03$ &  \scriptsize{($0.533$)} &  $0.05$ & \scriptsize{($0.195$)} \\
    \bottomrule
  \end{tabular}
\end{table}

Table~\ref{tab:pagerank} shows the correlations between plausibility and the difference in Page\-Rank as a proxy for the recognition heuristic. While, we have not obtained  statistically strong correlation in the datasets, for two of the datasets (\data{Quality} and \data{Traffic}) the direction of the correlation is according to the expectation: plausibility rises with increased recognition. 
More research to establish the degree of actual recognition and PageRank values is thus needed. 
Nevertheless, to our knowledge, this is the first experiment that attempted to use PageRank to model recognition.

\section{Semantic Coherence}
\label{sec:coherence}

\citet{ExplanatoryCoherence} has noted the importance of coherence for explanatory power. This concept is closely related to epistemic interpretability. Note, however, that it is not only important that the explanation is coherent with existing background knowledge but the explanatory factors should also be coherent with each other, as well as with the concept that should be explained. \citet{ExplanatoryCoherence} writes that ''\emph{a hypothesis coheres with propositions that it explains, or that explain it, or that participate with it in explaining other propositions, or that offer analogous explanations.}''

In previous work \citep{PaulheimESWC2012}, we conducted experiments with various statistical datasets enriched with Linked Open Data, one being the already mentioned \emph{Quality of Living} dataset, another one denoting the corruption perceptions index (CPI)\footnote{\url{https://www.transparency.org/research/cpi/overview}} in different countries worldwide. For each of those, we created rules and had them rated in a user study.

From that experiment, we experienced that many people tend to trust rules more if there is a high semantic coherence between the conditions in the rule. For example, a rule stating the the quality of living in a city is high if it is a \emph{European capital of culture} and is the headquarter of \emph{many book publishers} would be accepted since both conditions refer to cultural topics, whereas a rule involving \emph{European capital of culture} and many \emph{airlines founded} in that city would be considered to be less plausible.

Figure~\ref{fig:semantic-rules} depicts a set of results obtained on an unemployment statistic for French departments, enriched with data from DBpedia \cite{ristoski2013analyzing}. There are highly coherent rules combining attributes such as latitude and longitude, or population and area, as well as lowly coherent rules, combining geographic and demographic indicators. Interestingly, all those combinations perform a similar split of the dataset, i.e., into the continental and overseas departments of France.

\begin{figure}[t]
  \begin{Verbatim}
  
        Unemployment = low  :- area > 6720,
                               population > 607430.
        Unemployment = high :- latitude <= 44.1281, 
                               longitude <= 6.3333, longitude > 1.8397.
	\end{Verbatim}
	
		\centering (a) good discriminative rules, highly coherent
		
		\medskip

	\begin{Verbatim}
	
	Unemployment = high :- latitude <= 44.189,
                               population <= 635272 
	Unemployment = high :- longitude > 1.550,
                               population > 282277
        \end{Verbatim}
	
		\centering (b) good discriminative rules, lowly coherent
	
	\caption{Example rules for unemployment in different French regions}
	\label{fig:semantic-rules}
\end{figure}

At first glance, semantic coherence and discriminative power of a rule look like a contradiction, since semantically related attributes may also correlate: as in the example above, attributes describing the cultural life in a city can be assumed to correlate more strongly than, say, cultural and economic indicators. Hence, it is likely that a rule learner, without any further modifications, will produce semantically incoherent rules at a higher likelihood than semantically coherent ones.

However, in \citet{DM-NLP-14-CoherentRules}, we have shown that it is possible to modify rule learners in a way so that they produce more coherent rules. To that end, attribute labels are linked to a semantic resource such as WordNet \citep{wordnet}, and for each pair of attributes, we measure the distance in that semantic network. In the first place, this provides us with a measure for semantic coherence within a rule. Next, we can explicitly use that heuristic in the rule learner, and combine it with traditional heuristics that are used for adding conditions to a rule. Thereby, a rule learner can be modified to produce rules that are semantically coherent. 

\enlargethispage*{12pt}

The most interesting finding of the above work was that semantically coherent rules can be learned without significantly sacrificing accuracy of the overall rule-based model. This is possible in cases with lots of attributes that a rule learner can exploit for achieving a similar split of the dataset. In the above example with the French departments, any combination of latitude, longitude, population and area can be used to discriminate continental and overseas departments; therefore, the rule learner can pick a combination that has both a high discriminative power and a high coherence.

\section{Structure}
\label{sec:structure}

Another factor which, in our opinion, contributes strongly to the interpretability of a rule-based model is its internal logical structure. 
Rule learning algorithms typically provide flat lists that directly relate the input
to the output. Consider, e.g., the extreme case of learning a parity concept, which checks whether an odd or an even number of $r$ relevant attributes (out of a possibly higher total number of attributes) are set to \emph{true}.
Figure~\ref{fig:parity}(a) shows a flat rule-based representation of the target concept for $r = 5$, which requires $2^{r-1} = 16$ rules, whereas a structured representation, which introduces three auxiliary predicates (\texttt{parity2345}, \texttt{parity345}, and \texttt{parity45}) is much more concise using only $2\cdot(r-1) = 8$ rules (Figure~\ref{fig:parity}(b)). We argue that the parsimonious structure of the latter is much easier to comprehend because it uses only a linear number of rules, and slowly builds up the complex target concept \texttt{parity} from the smaller subconcepts \texttt{parity2345}, \texttt{parity345}, and \texttt{parity45}. 
This is in line with the criticism of \citet{FuzzyRules-DataDriven} who argued that the flat structure of fuzzy rules is one of the main limitations of current fuzzy rule learning systems.

\begin{figure}[t]
\begin{small}
	\begin{Verbatim}
	parity :-     x1,     x2,     x3,     x4, not x5.
	parity :-     x1,     x2, not x3, not x4, not x5.
	parity :-     x1, not x2,     x3, not x4, not x5.
	parity :-     x1, not x2, not x3,     x4, not x5.
	parity :- not x1,     x2, not x3,     x4, not x5.
	parity :- not x1,     x2,     x3, not x4, not x5.
	parity :- not x1, not x2,     x3,     x4, not x5.
	parity :- not x1, not x2, not x2, not x4, not x5.
	parity :-     x1,     x2,     x3, not x4,     x5.
	parity :-     x1,     x2, not x3,     x4,     x5.
	parity :-     x1, not x2,     x3,     x4,     x5.
	parity :- not x1,     x2,     x3,     x4,     x5.
	parity :- not x1, not x2, not x3,     x4,     x5.
	parity :- not x1, not x2,     x3, not x4,     x5.
	parity :- not x1,     x2, not x3, not x4,     x5.
	parity :-     x1, not x2, not x2, not x4,     x5.
	\end{Verbatim}
	
	\centering (a) flat unstructured rule set
	
	\medskip
	
	\begin{Verbatim}
	parity45   :-     x4,     x5.
	parity45   :- not x4, not x5.
	
	parity345  :-     x3, not parity45.
	parity345  :- not x3,     parity45.  
	
	parity2345 :-     x2, not parity345.
	parity2345 :- not x2,     parity345.   
	
	parity     :-     x1, not parity2345.
	parity     :- not x1,     parity2345. 
	\end{Verbatim}
	
\end{small}
	
	\centering (b) deep structured rule base with three invented predicates
	
	\caption{Unstructured and structured rule sets for the parity concept.}
	\label{fig:parity}
\end{figure}

However, we are not aware of psychological work that supports this hypothesis.
The results of a small empirical validation were recently reported by \citet{PI-Comprehensibility}, who performed a user study in which the participants were
shown differently structured elementary theories from logic programming, such as definitions for 
\texttt{grandfather}, \texttt{greatgrandfather}, or \texttt{ancestor}, and it was observed how quickly queries about a certain ancestry tree could be answered using these predicates.
Among others, the authors posed and partially confirmed the hypothesis that logical programs are more comprehensible if they are structured in a way that leads to a compression in length. In our opinion, further work is needed in order to see whether compression is indeed the determining factor here.
It also seems natural to assume that an important prerequisite for structured theories to be more comprehensible is that the intermediate concepts are by themselves meaningful to the user. Interestingly, this was not confirmed in the experiments by \citet{PI-Comprehensibility}, where the so-called ''public'' setting, in which all predicates had meaningful names, did not lead to consistently lower answer times than the ''private'' setting, in which the predicates did not have meaningful names.
They also could not confirm the hypothesis that it furthered comprehensibility when their participants were explicitly encouraged to think about meaningful names for intermediate concepts.

Although there are machine learning systems that can tackle simple problems like the family domain, there is no
system that is powerful enough to learn deeply structured logic theories for realistic problems, on which we could rely for experimentally testing this hypothesis. In machine learning, this line of work has been known as
{\em constructive induction\/} \cite{CI-Framework} or \emph{predicate invention}
\cite{ILP-PI},
but surprisingly, it has not received much attention since the classical works in inductive logic programming in the 1980s and 1990s.
One approach is to use a wrapper to scan for regularly co-occurring patterns in rules, and use them to define new intermediate concepts which allow to compress the original theory \cite{AQ17-HCI,CiPF,New-MDL}. 
Alternatively, one can directly
invoke so-called predicate invention operators during the learning process, as, e.g., in Duce \cite{Duce}, which operates in propositional logic, and its successor systems in first-order logic \cite{CIGOL,CHAMP,PI-Statistical}. One of the few recent works in this area is by \citet{ILP-PI-MetaInterpretative}, who introduced a technique that employs user-provided meta rules for proposing new predicates.

None of these works performed a systematic evaluation of the generated structured theories from the point of view of interpretability. Systems like MOBAL \cite{Mobal}, which not only tried to learn theories from data, but also provided functionalities for reformulating and restructuring the knowledge base \cite{Sommer-Diss}, have not received much attention in recent years. We believe that providing functionalities and support
for learning structured knowledge bases is crucial for the acceptance of learned models in complex domains. In a way, the recent success of deep neural networks needs to be carried over to the learning of deep logical structures. Recent work on so-called sum-product nets, which combine deep learning with graphical models and generate new concepts in their latent variables \cite{SPN-Interpretation}, may be viewed as a step into this direction.


    


\section{Conclusion}

The main goal of this paper was to motivate that interpretability of rules is an important topic that has received far too little serious attention in the literature. 
Its main contribution lies in highlighting that plausibility is an important aspect of interpretability, which, to our knowledge, has not been investigated before.
In particular, we observed that even rules that have the same predictive quality in terms of conventional measures such as support and confidence, and will thus be considered as equally good explanations by conventional rule learning algorithms, may be perceived with different degrees of plausibility.

More concretely, we reported on five experiments conducted in order to gain first insight into plausibility of rule learning results. Users were confronted with pairs of learned rules with approximately the same discriminative power (as measured by conventional heuristics such as support and confidence), and were asked to indicate which one seemed more plausible. The experiments were performed in four domains, which were selected so that participants can be expected to be able to comprehend the given explanations (rules), but not to reliably judge their validity without obtaining additional information. In this way, users were guided to give an intuitive assessment of the plausibility of the provided explanation.

Experiment 1 explored the hypothesis whether the Occam's razor principle holds for the plausibility of rules, by investigating whether people consider shorter rules  to be more plausible than longer rules.  The results obtained for four different domains indicated that this might not be the case,  in fact we observed statistically significant preference for longer rules on two datasets.
In Experiment 2, we found support for the hypothesis that the elevated preference for longer rules is partly due to the misunderstanding of ``and'' that connects conditions in the presented rules: people erroneously find rules with more conditions as more general.  In Experiment 3, we focused on another ingredient of rules: the values of confidence and support metrics.  The results suggest that when both confidence and support are stated, confidence positively affects plausibility and support is largely ignored. This confirms a prediction following from previous psychological research studying the \emph{insensitivity to sample size} effect.
As a precursor to a follow-up study focusing  on the weak evidence effect, Experiment 4 evaluated the relation between perceived plausibility and strength of conditions in the rule antecedent. The results indicate that rule plausibility is affected already if a single condition is considered to be relevant. 
\emph{Recognition} is a powerful principle underlying many human reasoning patterns and biases. In Experiment 5, we attempted to use PageRank computed from Wikipedia graph as a proxy for how well a given condition is recognized.  The results, albeit statistically insignificant, suggest the expected pattern of positive correlation between recognition and plausibility. This experiment is predominantly interesting from the methodological perspective, as it offers a possible approach to approximation of recognition of rule conditions.

We acknowledge several limitations of the presented experiments. In particular, some of the results might be influenced by the specific domains of the datasets involved. For some of the experiments (Experiment 4) the collected number of judgments was also rather small affecting their statistical significance. Another limitation is the absence of expert users as we relied solely on judgments elicited with crowdsourcing.  Lastly, we lacked counsel of a psychologist skilled in designing and evaluating user experiments. Overall, we suggest our experimental results should be replicated on other domains addressing also the other limitations noted above.

%

%
In our view, a research program that aims at a thorough investigation of interpretability in machine learning needs to resort to results in the psychological literature, in particular to cognitive biases and fallacies.
We summarized some of these hypotheses, such as the conjunctive fallacy, and started to investigate to what extent these can serve as explanations for human preferences between different learned hypotheses.  There are numerous other cognitive effects that can demonstrate how people assess rule plausibility, some of which are briefly listed in Appendix~\ref{ss:otherheuristics} and discussed more extensively in \citet{kliegr2018review}.  Clearly, more work along these lines is needed.

Moreover, it needs to be considered how cognitive biases can be incorporated into machine learning algorithms. Unlike loss functions, which can be evaluated on data, it seems necessary that interpretability is evaluated in user studies. Thus, we need to establish appropriate evaluation procedures for interpretability, and develop appropriate heuristic surrogate functions that can be quickly evaluated and be optimized in learning algorithms. 

Finally, in our work we have largely ignored the issue of background knowledge
by picking domains in which we assumed that our participants have a basic knowledge
that allows them to judge the plausibility of rules. However, justifiability, i.e.,
whether a model is in line with existing background knowledge, is an important prerequisite for plausibility. The work of \citet{justifiability} is pioneering
in that they try to formalize this notion in the context of domain knowledge.
Based on this, another promising research direction is infusing semantic metadata into the learning process and exploiting it for enforcing the output of rules that are likely to be accepted more by the end user.

\bigskip

\vfill

  \begin{small}
  	\noindent
\textbf{Acknowledgments.}
We would like to thank Frederik Janssen and Julius Stecher for providing us with their code, Eyke H\"ullermeier, Frank J\"akel,
Niklas Lavesson, Nada Lavra\v{c} and Kai-Ming Ting for interesting discussions and pointers to related work,
and
Jilles Vreeken for 
pointing us to \citet{KolmogorovDirections}.
We are also grateful for the insightful comments of the anonymous reviewers, which helped
us considerably to focus our paper, and provided us with many additional pointers to relevant works in the literature.
TK was supported by grant IGA 33/2018 of the Faculty of Informatics and Statistics, University of Economics, Prague.
 
\end{small}

\newpage

\bibliographystyle{mynatbib}

\newpage 

\appendix

%
%
%
%

\section*{Appendix -- A Brief Overview of Relevant Cognitive Heuristics and Fallacies}
\label{ss:otherheuristics}

%
In this appendix we provide a list of cognitive phenomena that can be important for interpretation of rule learning results. 
However, we neither claim completeness, nor can we provide 
more than a very short summary of each phenomenon. A more extensive
treatment can be found in \citep{kliegr2018review}.
 An extensive
treatment of the subject can be found in \citep{kliegr2018review}.

The list is divided in three categories. The first two cover cognitive biases (also called illusions) that are included in a recent authoritative review by \citet{pohl2017cognitive}. The first category, \emph{Thinking}, covers those related to thinking processes. These require the person to apply a certain rule (such as the Bayes theorem). Since many people do not know this rule, they have to apply it intuitively, which can result in errors. The second category, \emph{Judgment}, covers biases used by people when they are asked to rate some property of a given object (such as a plausibility of a rule). Note that \citet{pohl2017cognitive} also defined a third category, \emph{Memory}, which we do not consider as directly relevant to our problem. Instead, we introduce ``\emph{Other}'' category into which we put cognitive phenomena that were not explicitly systematized by \citet{pohl2017cognitive}, although many of the phenomena listed under it clearly belong to one of the established categories.

\paragraph{Thinking.}
\begin{itemize}
\item \emph{Base rate neglect} \citep{1974-02325-00119730701,bar1980base}.  Insensitivity to the prior probability of the outcome, violating the principles of probabilistic reasoning, especially Bayes' theorem.

\item \emph{Confirmation bias and positive test strategy} \citep{nickerson1998confirmation}. Seeking  or  interpretation   of  evidence  so that it conforms to  existing  beliefs, expectations, or a hypothesis  in hand.

\item \emph{Conjunction fallacy and representativeness heuristic} \citep{tversky1983extensional}. Conjunction fallacy occurs  when a person assumes that a specific condition is more probable than a single general condition in case the specific condition seems as more representative of the problem at hand.

\end{itemize}

\enlargethispage*{10pt}

\paragraph{Judgment.}

\begin{itemize}
\item \emph{Availability heuristic} \citep{tversky1973availability}. The easier it is to recall a piece of information, the greater the importance of the information.

\item \emph{Effect of difficulty} \citep{griffin1992weighing}. If it is difficult to tell which one of two mutually exclusive alternative hypotheses is better because both are nearly equally probable, people will grossly overestimate the confidence associated with their choice. This effect is also sometimes referred to as \emph{overconfidence effect} \citep{pohl2017cognitive}.

\item \emph{Mere-exposure effect} \citep{zajonc1968attitudinal}. Repeated encounter of a hypothesis results in increased preference.

\end{itemize}

\paragraph{Other.}

\begin{itemize}
\item \emph{Ambiguity aversion} \citep{ellsberg1961risk}. People tend to favour options for which the probability of a favourable outcome is known over options where the probability of favourable outcome is unknown. Some evidence suggests that ambiguity aversion has a genetic basis \citep{Chew2012}.

\item \emph{Averaging heuristic} \citep{fantino1997conjunction}. Joint probability of two events is estimated as an average of probabilities of the  component events. This fallacy corresponds to believing that $P(A,B) = \frac{P(A) + P(B)}{2}$ instead of $P(A,B) = P(A) * P(B)$.

\item \emph{Confusion of the inverse} \citep{plous1993psychology}. Conditional probability is equivocated with its inverse. This fallacy corresponds to believing that $P(A|B) = P(B|A)$.

\item \emph{Context and trade-off contrast} \citep{tversky1993context}. The tendency to prefer alternative $x$ over alternative $y$ is influenced by the context -- other available alternatives.  

\item \emph{Disjunction fallacy} \citep{bar1993alike}. People tend to think that it is more likely for an object to belong to a more characteristic subgroup than to its supergroup.

\item \emph{Information bias} \citep{baron1988heuristics}.
People tend to belief that more information the better, even if the extra information is irrelevant for their decision.
 
\item \emph{Insensitivity to sample size} \citep{Tversky27091974}.  Neglect of the following two principles: a) more variance is likely to occur in smaller samples, b) larger samples provide less variance and better evidence.

\item \emph{Recognition heuristic} \citep{goldstein1999recognition}. If  one  of  two  objects  is  recognized  and  the other is not, then infer that the recognized object has the higher value with respect to the criterion.

\item \emph{Negativity bias} \citep{kanouse1987negativity}. People weigh negative aspects of an object more heavily than positive ones.

\item \emph{Primacy effect} \citep{thorndike1927influence}. This effect can be characterized by words of Edward Thorndike (1874-1949), one of the founders of modern education psychology, as follows: ``other things being equal the association first formed will prevail'' \citep{thorndike1927influence}.

\item \emph{Reiteration effect} \citep{hasher1977frequency}. Frequency of occurrence is a criterion used to establish validity of a statement.

\enlargethispage*{10pt}


\item \emph{Unit bias} \citep{geier2006unit}. People  tend to give equal weight to each condition at the expense of detailed scrutiny of its actual weight.

\item \emph{Weak evidence effect} \citep{fernbach2011good}. Presenting weak, but supportive evidence makes people less confident in predicting a particular outcome than presenting no evidence at all.

\end{itemize}

\bigskip
\noindent
While this list is certainly not exhaustive, it is long enough to illustrate that 
interpretability is a very complex research challenge that cannot be met in passing but needs serious attention in our research programs.




\end{document}